\documentclass[conference]{IEEEtran}
\def\BibTeX{{\rm B\kern-.05em{\sc i\kern-.025em b}\kern-.08em
    T\kern-.1667em\lower.7ex\hbox{E}\kern-.125emX}}

\usepackage{ellipsis, ragged2e} 
\usepackage[l2tabu, orthodox]{nag} 

\usepackage[english]{babel}
\usepackage[T1]{fontenc}
\usepackage[utf8]{inputenc}

\usepackage{amsmath,amsfonts,amssymb}
\usepackage{xcolor,graphicx,wrapfig,paralist}
\usepackage{booktabs}
\usepackage{tikz}
\usepackage{csquotes}
\usepackage{graphics}
\usepackage{stmaryrd}
\usepackage{url}
\usepackage{bbm}
\usepackage{textcomp}
\usepackage{xcolor}

\usepackage{amsthm}
\usepackage{thmtools}
\usepackage{thm-restate}

\usepackage{pgfplots}

\usepackage{tabu,multirow}
\usepackage{xparse}
\usepackage{mathtools}
\usepackage{xfrac}

\usepackage{caption}
\usepackage{subcaption}

\usepackage{algorithmicx,algorithm}
\usepackage[noend]{algpseudocode}

\usepackage[capitalize]{cleveref}

\newtheorem{theorem}{Theorem}

\newtheorem{lemma}{Lemma}

\theoremstyle{definition}
\newtheorem{definition}{Definition}
\newtheorem{remark}{Remark}
\newtheorem{example}{Example}

\usetikzlibrary{shapes,positioning,arrows,arrows.meta,calc,automata,matrix,fit,backgrounds,fadings}

\usepackage{tikzsymbols}
\pgfplotsset{compat=1.14} 

\tikzstyle{maxstate}=[fill=white,draw,rectangle,inner sep=3pt,rounded corners=0pt,minimum size=7mm]
\tikzstyle{minstate}=[fill=white,draw,circle,inner sep=2pt,minimum size=8mm]
\tikzstyle{action}=[font=\small,inner sep=3pt,anchor=center,outer sep=2pt,fill=white,text opacity=1,fill opacity=0.8,scope fading=circle with fuzzy edge 20 percent]
\tikzstyle{loopaction}=[font=\small,outer sep=2pt]
\tikzstyle{actionnode}=[circle,draw=black,fill=black,minimum size=1mm,inner sep=0,outer sep=0]
\tikzstyle{actionedge}=[draw,-]
\tikzstyle{prob}=[font=\scriptsize,inner sep=3pt,outer sep=0pt,anchor=center,color=black!80,fill=white,text opacity=1,fill opacity=0.8,scope fading=circle with fuzzy edge 20 percent]
\tikzstyle{probedge}=[draw,->]
\tikzstyle{directedge}=[draw,->]

\tikzset{chainarrow/.tip={Stealth[length=3pt]}}
\tikzset{>=chainarrow}
\renewcommand{\Box}{\square}

\newcommand{\unionSym}{\cup}
\newcommand{\unionBin}{\mathbin{\unionSym}}

\newcommand{\intersectionSym}{\cap}
\newcommand{\intersectionBin}{\mathbin{\intersectionSym}}
\newcommand{\UnionSym}{\bigcup}

\newcommand{\union}{\unionBin}

\newcommand{\intersection}{\intersectionBin}
\newcommand{\Union}{\UnionSym}

\DeclareGraphicsExtensions{.pdf, .png, .jpg}

\newcommand{\abs}[1]{\lvert #1 \rvert}

\newcommand{\expec}{\mathbb{E}}

\newcommand{\Naturals}{\mathbb{N}}
\newcommand{\Reals}{\mathbb{R}}
\newcommand{\RealsNonneg}{\mathbb{R}_{\geq 0}}

\DeclareMathOperator*{\argopt}{arg\, opt}
\DeclareDocumentCommand{\post}{D<>{} O{} D(){}}{\mathsf{Post}_{#1}^{#2}\IfValueT{#3}{(#3)}}
\newcommand{\eqdef}{\vcentcolon=}

\newcommand{\drawbox}{\node[draw,rectangle,minimum size=.7cm, outer sep=1pt]}
\newcommand{\drawdummy}{\node[minimum size=0,inner sep=0]}

\newcommand{\MC}{\mathsf{M}}
\newcommand{\MDP}{\mathcal{M}}
\newcommand{\infinitepath}{\rho}
\NewDocumentCommand{\Infinitepaths}{d<>}{\IfNoValueTF{#1}{\mathsf{Paths}}{\mathsf{Paths}_{#1}}}
\renewcommand{\path}{\infinitepath}

\newcommand{\distribution}{d}
\NewDocumentCommand{\Distributions}{d()}{\IfNoValueTF{#1}{\mathcal{D}}{\mathcal{D}(#1)}}

\newcommand{\G}{\mathsf{G}}
\newcommand{\allstates}{\mathsf{S}}
\newcommand{\states}{\allstates}
\newcommand{\maxstates}{\allstates_{\Box}}
\renewcommand{\circ}{\bigcirc}
\newcommand{\minstates}{\allstates_{\circ}}
\newcommand{\initstate}{s_{0}}
\newcommand{\act}{\mathsf{A}}
\newcommand{\actions}{\act}

\newcommand{\trans}{\delta}

\newcommand{\targets}{\mathsf{F}}
\newcommand{\fstates}{\targets}

\newcommand{\statereward}{\mathsf{r}}
\newcommand{\fpoff}{\mathsf{off}}
\newcommand{\meanpay}{\mathit{mp}}

\newcommand{\reach}{\lozenge}
\newcommand{\straa}{\sigma}
\newcommand{\strab}{\tau}

\newcommand{\probability}{\mathbb{P}}

\newcommand{\val}{\mathsf{V}}
\newcommand{\lb}{\mathsf{L}}
\newcommand{\ub}{\mathsf{U}}

\newcommand{\opt}{\mathsf{opt}}
\newcommand{\stayVal}{\mathsf{stay}}
\newcommand{\upperStayVal}{\mathsf{U}\stayVal}
\newcommand{\lowerStayVal}{\mathsf{L}\stayVal}
\newcommand{\bE}{\mathsf{exit}}
\DeclareMathOperator{\exits}{\mathsf{exits}}
\newcommand{\restrict}{\mid E}

\newcommand{\aIndent}{\hspace{\algorithmicindent}}

\newcommand{\initViVal}{\textsc{InitVI}_\Phi}
\newcommand{\upperViVal}{\textsc{InitU}_\Phi}

\newcommand{\fixpointup}{\mathsf{Fix}}

\usepackage{etoolbox}
\newtoggle{arxiv}

\toggletrue{arxiv} 

\newcommand{\appref}[1]{Appendix~\ref{#1}}

\newcommand{\qee}{\hfill$\triangle$} 

\newcommand{\commsym}{\triangleright}
\renewcommand{\commsym}{\#}
\algrenewcommand\algorithmiccomment[1]{\hfill\emph{$\commsym$\ #1}}
\newcommand{\Commentline}[1]{\emph{$\commsym$\ #1}}

\IEEEoverridecommandlockouts

\iftoggle{arxiv}{}{
\IEEEpubid{\makebox[\columnwidth]{979-8-3503-3587-3/23/\$31.00~
\copyright2023 IEEE \hfill} \hspace{\columnsep}\makebox[\columnwidth]{ }}

\setlength{\textfloatsep}{0.25\textfloatsep} 
\setlength{\intextsep}{0.25\intextsep} 
\setlength{\floatsep}{0.25\floatsep} 
\setlength{\abovecaptionskip}{0.75\abovecaptionskip}
\setlength{\belowcaptionskip}{0.75\belowcaptionskip}

\renewcommand{\baselinestretch}{0.974} 
}

\begin{document}

\title{Stopping Criteria for Value Iteration on Stochastic~Games with Quantitative Objectives
	\thanks{This research was funded in part by DFG projects 383882557 \enquote{SUV} and 427755713 \enquote{GOPro}.}
}
\author{
	\hfill \begin{tabular}{ccc}
		\begin{minipage}{0.3\textwidth}
			\centering
			Jan K\v{r}et{\'i}nsk{\'y}\IEEEauthorrefmark{3}\IEEEauthorrefmark{2} \\
			0000-0002-8122-2881
		\end{minipage} &
		\begin{minipage}{0.3\textwidth}
			\centering
			Tobias Meggendorfer\IEEEauthorrefmark{3}\IEEEauthorrefmark{1}
			0000-0002-1712-2165
		\end{minipage} &
		\begin{minipage}{0.3\textwidth}
			\centering
			Maximilian Weininger\IEEEauthorrefmark{3}\IEEEauthorrefmark{1}
			0000-0002-0163-2152
		\end{minipage}
	\end{tabular} \hfill \\[1.5ex]

	\hfill
	\begin{tabular}{ccc}
		\begin{minipage}{0.4\textwidth}
			\centering
			\IEEEauthorrefmark{1}\textit{Institute of Science and Technology Austria}\\
			Klosterneuburg, Austria
		\end{minipage} &
		\begin{minipage}{0.225\textwidth}
			\centering
			\IEEEauthorrefmark{2}\textit{Masaryk University}\\
			Brno, Czech Republic
		\end{minipage} &
		\begin{minipage}{0.3\textwidth}
			\centering
			\IEEEauthorrefmark{3}\textit{Technical University of Munich}\\
			Garching bei München, Germany
		\end{minipage}
	\end{tabular}
	\hfill
	\vspace{-1em}
}


\maketitle

\begin{abstract}
A classic solution technique for Markov decision processes (MDP) and stochastic games (SG) is value iteration (VI).
Due to its good practical performance, this approximative approach is typically preferred over exact techniques, even though no practical bounds on the imprecision of the result could be given until recently.
As a consequence, even the most used model checkers could return arbitrarily wrong results.
Over the past decade, different works derived stopping criteria, indicating when the precision reaches the desired level, for various settings, in particular MDP with reachability, total reward, and mean payoff, and SG with reachability.

In this paper, we provide the first stopping criteria for VI on SG with total reward
and mean payoff, yielding the first anytime algorithms in these settings.
To this end, we provide the solution in two flavours:
First through a reduction to the MDP case and second directly on SG.
The former is simpler and automatically utilizes any advances on MDP.
The latter allows for more local computations, heading towards better practical efficiency.

Our solution unifies the previously mentioned approaches for MDP and SG and their underlying ideas.
To achieve this, we isolate objective-specific subroutines as well as identify objective-independent concepts.
These structural concepts, while surprisingly simple, form the very essence of the unified solution.


\end{abstract}

\begin{IEEEkeywords}
	Stochastic games, value iteration
\end{IEEEkeywords}

\section{Introduction}

\newcommand{\valintro}{v^*}

\textbf{Markov decision processes (MDP) and stochastic games (SG)} are the standard basic models for sequential decision making in the presence of uncertainty.
While MDP can model either controllable behaviour or adversarial behaviour (such as schedulers or unknown environment), SG generalize MDP to encompass both at once.
A \emph{quantitative objective} for such a system is a random variable, assigning a real number to each run.
A classic task is to find the optimal value $\valintro$ of its expectation, maximized (or minimized) over all strategies (a.k.a.\ policies, schedulers, or controllers) resolving the choices.
For decades, this problem has been thoroughly investigated for both formalisms with respect to various {quantitative objectives}, such as discounted reward \cite{shapley}, total reward \cite{TV87}, long-run average reward (a.k.a. mean payoff) \cite{gillette1957stochastic,NM81}, or reachability/safety \cite{condonAlgo} as an important special case of the latter two.

\textbf{Value iteration (VI)} \cite{bellman} is a classic solution technique applicable to these settings \cite{DBLP:conf/spin/ChatterjeeH08}, yielding a sequence of approximants converging in the limit to the actual optimal value $\valintro$.
It often is preferred over other solution methods due to its good practical performance, see \cite{unpublished:mdplp} for a comparison on MDP.
For the use in applications such as verification of safety-critical systems, guarantees on the \emph{precision of the approximants} play a paramount role since the exact optimal value $\valintro$ may not be attained by any of the approximants within the given computation-time budget or in any finite time.
Besides, it is typically sufficient to determine $\valintro$ with a specified precision.
In this case an earlier termination of the algorithm is desirable, while still guaranteeing at least the required precision.

However, until recently, VI algorithms for non-discounted rewards provided no precision of the produced approximants and implementations in model checkers could yield arbitrarily wrong results \cite{HM18}!
Fortunately, recent works have equipped VI in several settings with \emph{stopping criteria}, i.e.\ the algorithms produce not only approximants of $\valintro$ but also a bound on their imprecision, allowing VI to stop when the desired precision is reached.
In other words, the algorithms produce guaranteed lower and upper bounds on $\valintro$.
Since these under- and over-approximations of $\valintro$ are steadily improving and accessible at any point of time during the (possibly infinite) run of the VI algorithm, VI becomes a so-called \emph{anytime algorithm}, i.e.\ at every step in the algorithm, it can return the current estimate with its imprecision/error bound, and this bound converges to 0 in the limit.
This has been achieved for MDP with various objectives such as reachability \cite{atva,HM18}, total reward until reaching the target \cite{ensuring,DBLP:conf/cav/QuatmannK18,DBLP:conf/cav/HartmannsK20}, long-run average reward \cite{DBLP:conf/cav/AshokCDKM17}, but for SG due to their more complex structure only for reachability \cite{KKKW18,widestPaths} and multi-dimensional reachability \cite{DBLP:conf/lics/AshokCKWW20}.
As emphasized in the conclusion of \cite{prism-games}, these shortcomings are still to be studied for SG.

\textbf{Our contribution} in this paper are the \emph{first stopping criteria} for total reward and long-run average reward in SG.
This enables a reliable use of VI, as required e.g.\ in model checkers.
Moreover, our approach is a \emph{unified solution} for the various objectives and systems. 
To better identify the commonalities of the various objectives and achieve deeper understanding, we provide the solution in two flavours.

The first algorithm reduces the problem on SG to a sequence of problems on MDP.
These can be solved by any algorithm for MDP. 
Consequently, this approach directly profits from any improvements on solving MDP.
Moreover, our algorithm generalizes to a stopping criterion for strategy iteration (SI)~\cite{howard} in SG:
While SI classically is required to be executed until the optimum is obtained in order to yield guarantees, we obtain converging bounds and thus an anytime algorithm, too.

The second algorithm extends the guaranteed-precision VI from MDP to SG.
It uses the bounded variant of VI \cite{brtdp}, where VI iterates both from below and separately from above, thus providing its own error bounds.
Since this algorithm operates \emph{directly on SG}, it can localize the convergence difficulties (we even give a characterization of when this happens) and overcome them by local and thus possibly cheaper computation.

On the conceptual level, our work can be seen as an extension of previous work from \emph{two perspectives}: 

Firstly, more visibly in the first algorithm, we \emph{extend the traditional VI} algorithms for the various objectives in SG to VI \emph{with a stopping criterion}.
Instead of computing one sequence converging to $\valintro$, we compute two sequences monotonically converging from below and from above, forming the under- and over-approximations, respectively.
Interestingly, depending on the objective, traditional VI algorithms may be reused to different extents.
For instance, the traditional VI for reachability already yields a valid under-approximation and we only have to complement it with a valid over-approximation.
Dually, for safety, the traditional VI provides an over-approximation to be complemented by an under-approximation.
In contrast, for long-run average reward -- a generalization of both reachability of safety -- new approximations are needed on both sides since the original VI sequence is oscillating around the value $\valintro$.

Secondly, more visibly in the second algorithm, we \emph{extend the VI stopping criteria} for MDP and for reachability in SG \emph{to SG with more complex objectives}.
VI essentially computes a fixpoint of an update rule for the value in each states.
The issue in getting both converging under- and over-approximation is that there may be more fixpoints because of cyclic dependencies of the values.
In MDP, this takes the form of all states in a so-called \emph{maximal end component} (MEC) having the same value; the solution is thus simply grouping all these states into a single one (often called MEC-quotient or collapsing MECs \cite{atva,HM18}).
While these MECs can be identified on the graph-theoretical level, in SG the solution requires a finer concept of so-called \emph{simple end components} (SEC) \cite{KKKW18}, which depend on the values that we are in turn trying to compute (as explained in \cref{ex:changing_secs_in_games}).
This vicious cycle is broken by considering varying sets of states to be grouped only temporarily during the run of the algorithm, depending on the current approximations.
Finally, in contrast to reachability, where remaining in such sets of states yields only values 0 or 1, yet another level of difficulty in grouping arises for total and average reward, where non-trivial values can be attained. 

Our contribution can be thus summarized as follows:
\begin{itemize}
	\item We give the first stopping criteria for VI on SG with total reward (in its three common variants \cite{SGreward}) and with long-run average reward, hence the first reliable VI algorithms. 
	Moreover, they are in the form of anytime algorithms.
	\item Our approach provides a uniform framework for the new as well as previously studied cases with the standard quantitative objectives.
	In the course of this, we identify fundamental objective-independent concepts, such as \enquote{staying} and \enquote{exiting} values, together with the
objective-specific ways to compute them, or a new definition of SEC.
	The new concepts apply uniformly to all considered objectives.
	Consequently, our solution is also more transparent.
	Finally, due to its generality, it yields a stopping criterion for strategy iteration, too.
\end{itemize}

%
%

\subsection{Further Related Work}

Beside VI there are several techniques used to solve MDP or SG.
\emph{Linear programming (LP)} typically yields solutions to MDP problems running in polynomial time.
In contrast, for games, LP cannot capture the opposite goals of the players and we have to resort to quadratic programming~\cite{condonAlgo}, which, using current methods, is either practically infeasible or exponential in the size of the SG~\cite{gandalf}.
Besides, even in MDP the use of LP is practically limited, since it does not perform well on larger models~\cite{atva17,unpublished:mdplp}.
Thus, for both MDP and SG, usually either strategy iteration (SI) or value iteration are preferred.


\emph{Strategy iteration} \cite{howard} computes a sequence of improving strategies in at most exponentially many iterations \cite{DBLP:conf/lics/Friedmann09,DBLP:journals/corr/abs-1106-0778}.
Upon termination, the algorithm returns a guaranteed optimal strategy, i.e.\ one achieving the optimal value.
However, the intermediate results only provide one-sided (converging) approximation of the result.
(Indeed, SI improves and evaluates the strategies of one player only, since alternating improvement for both players may not terminate due to cycling~\cite{condonAlgo}.)
Consequently, in order to obtain a guaranteed approximation, SI must finish the whole computation, which may be infeasible for large systems.
Our stopping criterion extends to SI, providing a simple fix, and thus yields (to the best of our knowledge) the first anytime SI algorithm with guaranteed approximation even before termination.
%

\emph{Value iteration} only is guaranteed to yield the precise result after a number of steps exponential in the size of the SG as well as the denominators of its quantities \cite{DBLP:conf/spin/ChatterjeeH08}, which is infeasible even for toy games.
Moreover, there are SG (even MDP) where this exponential number of steps indeed is necessary~\cite{DBLP:conf/icalp/BalajiK0PS19}.
Despite the imprecision for the practically whole time of the algorithm execution, using VI to compute $\varepsilon$-precise solutions has been the most popular technique for MDP \cite{prism,storm} and SG \cite{prism-games} even without stopping criteria.
This is because it (i)~performs even better than SI on MDP \cite{atva17,unpublished:mdplp}, (ii)~uses heuristics exploiting that the result is imprecise anyway (asynchronous VI and partial exploration \cite{brtdp,atva,DBLP:conf/atva/Meggendorfer22}), and (iii)~performs on par with SI on SG \cite{gandalf}.

The survey \cite{DBLP:conf/spin/ChatterjeeH08} provides a general treatment of VI on various models and objectives.
While it discusses convergence and the worst-case analysis, we strengthen this with error bounds and anytime algorithms, providing better practical performance.
It also conjectures VI is extensible to SG with long-run average reward, which we prove in this paper.
For total reward, our starting point are algorithms presented in \cite{SGreward} which converge in the limit.

Finally, there are also further, mostly theoretical techniques to solve games, such as reducing SG with long-run average reward to SG with discounted reward or with reachability \cite{DBLP:conf/isaac/AnderssonM09} at the cost of introducing impractically small quantities, leading to a very poor convergence rate.

%
%
%
%

\section{Preliminaries}

In this section, we recall the basics of (turn-based) stochastic games, introduce relevant objectives and sketch existing solution approaches.

We write $\Reals$ and $\RealsNonneg$ to denote the real numbers and the non-negative real numbers, respectively. 
For a set $S$, $S^\star$ and $S^\omega$ refer to the set of finite and infinite sequences of elements of $S$, respectively.
Whenever comparing two vectors $x$ and $y$ by $x \leq y$ or taking their $\max$ or $\min$, it is done point-wise.
$\Distributions(X)$ denotes the set of all \emph{probability distributions} over a countable set $X$, i.e.\ mappings $\distribution : X \to [0, 1]$ such that $\sum_{x \in X} \distribution(x) = 1$.
We also define $\max \emptyset = -\infty$ and $\min \emptyset = \infty$.
%
%

We refer to, e.g., \cite{DBLP:books/wi/Puterman94,DBLP:books/daglib/0020348,DBLP:conf/sfm/ForejtKNP11,filar2012competitive} for further information related to the topics discussed in the following.


\subsection{Stochastic Systems}
%
%
A \emph{Markov chain (MC)} (e.g.\ \cite{DBLP:books/daglib/0020348}), is a tuple $\MC = (\allstates, \trans)$, where
$\allstates$ is a finite set of \emph{states}, and
$\trans : \allstates \to \Distributions(\allstates)$ is a \emph{transition function} that for each state $s$ yields a probability distribution over successor states.

A \emph{(turn-based) stochastic game (SG)} (e.g.\ \cite{condonAlgo}) is a tuple $(\maxstates, \minstates, \act, \trans)$, where
$\maxstates$ and $\minstates$ are finite, disjoint sets of states belonging to the \emph{Maximizer} player $\Box$ and the \emph{Minimizer} player $\circ$, respectively, and induce $\allstates \eqdef \maxstates \union \minstates$;
further, $\act$ denotes a finite set of \emph{actions} and we overload $\act$ to also act as a function assigning to each state $s$ a set of non-empty \emph{available actions} $\act(s)$; and
$\trans : \allstates \times \act \to \Distributions(\allstates)$ is the \emph{transition function} that for each state $s$ and (available) action $a \in \act(s)$ yields a distribution over successor states.

A \emph{Markov decision process (MDP)} (e.g.\ \cite{DBLP:books/wi/Puterman94}) is an SG with only one player, i.e., $\maxstates = \emptyset$ or $\minstates = \emptyset$.

For convenience, we write $\trans(s, s')$ and $\trans(s, a, s')$ instead of $\trans(s)(s')$ and $\trans(s, a)(s')$, respectively.
Given a \emph{state-action pair} $s \in \allstates$, $a \in \act(s)$, we use $\post(s, a) \eqdef \{s' \in \allstates \mid \trans(s, a, s') > 0\}$ to denote the set of all successors of $s$ under action $a$.
For a state $s$, set ${\preccurlyeq^s} = {\leq}$ if $s \in \maxstates$ and $\geq$ otherwise. Analogously, $\opt^s$ is the $\max$-operator for $s \in \maxstates$ and the $\min$-operator otherwise, i.e.\ the preference of either player.
We omit the superscript $s$ when its clear from context.
For a function $f : \allstates \to \Reals$, let $f(s, a) \eqdef \sum_{s' \in \allstates} \trans(s, a, s') \cdot f(s')$ denote the expected value of $f$ achieved by following the action $a$ once.

\subsubsection*{Semantics}
We resolve choices by strategies, inducing a Markov chain with the respective probability space over infinite paths, as follows.
Intuitively, a stochastic game is played in turns:
In every state $s$, the player to whom it belongs chooses an action $a$ from the set of available actions $\act(s)$ and the play advances to a successor state $s'$ according to the probability distribution given by $\trans(s, a)$.
Starting in a state $s_0$ and repeating this process indefinitely yields an infinite sequence $\infinitepath = s_0 a_0 s_1 a_1 \dots \in (\allstates \times \act)^\omega$ such that for every $i \in \Naturals_0$ we have $a_i \in \act(s_i)$ and $s_{i+1} \in \post(s_i, a_i)$.
We refer to such sequences as \emph{(infinite) paths} or \emph{plays} and use $\Infinitepaths<\G>$ to denote the set of all such infinite paths in a given game $\G$.
Furthermore, we write $\infinitepath_i$ to denote the $i$-th state $s_i$ in a path.
\emph{Finite paths} or \emph{histories} are finite prefixes of a play, i.e.\ elements of $(\allstates \times \act)^\star \times \allstates$ consistent with $\act$ and $\trans$.

The path obtained by playing the game starting in a state $s$ both depends on the choices of the two players as well as the random outcomes.
The decision-making of the players is captured by the notion of \emph{strategies}.
In general, strategies are functions mapping a given history to a distribution over the actions available in the current state.
However, in our case, \emph{memoryless deterministic} strategies (abbreviated \emph{MD strategies}),
which choose a single action in each state irrespective of the history, are sufficient, as we argue later on.
Thus we immediately restrict ourselves to this simpler case.
Formally, a (MD) strategy of Maximizer $\straa : \maxstates \to \act$ or Minimizer $\strab : \minstates \to \act$ is a function mapping all states of the player to an available action, i.e.\ $\straa(s) \in \act(s)$ for all $s$.
A pair of strategies $\pi = (\straa, \strab)$ is called \emph{strategy profile}, and we define $\pi(s) \eqdef \straa(s)$ if $s \in \maxstates$ and $\strab(s)$ otherwise.

After fixing one player's strategy, the outcome only depends on the choices of the other player and the randomness -- this corresponds to an MDP.
Analogously, once both players chose a strategy, the behaviour of the system is defined by a Markov chain.
Given an appropriate pair of strategies $(\straa, \strab)$ for a game $\G$, we write $\G^{\straa, \cdot}$ and $\G^{\cdot, \strab}$ for the MDP obtained after fixing either Maximizer's or Minimizer's strategy, respectively.
Similarly, for a strategy profile $\pi = (\straa, \strab)$, we write $\G^\pi = \G^{\straa, \strab} = (\allstates, \hat{\trans})$, a Markov chain where $\hat{\trans}(s, s') \eqdef \trans(s, \pi(s), s')$.

Together with a state $s$, the Markov chain $\G^{\straa, \strab}$ induces a unique probability distribution $\probability_{\G, s}^{\straa, \strab}$ over the set of all infinite paths $\Infinitepaths<\G>$ \cite[Sec.~10.1]{DBLP:books/daglib/0020348} (where the set of paths starting in $s$ has measure 1).
For a random variable over paths $X : \Infinitepaths<\G> \to \Reals$, we write $\expec_{\G,s}^{\straa,\strab}[X]$ for the expected value of $X$ under the probability measure $\probability_{\G, s}^{\straa, \strab}$.

\subsection{Objectives}

Objectives formalize the \enquote{goal} of both players by assigning a value to each path.
The two players are antagonistic, and, as their names suggest, Maximizer aims to maximize the obtained value, while Minimizer wants to minimize it (i.e.\ the game is zero-sum).
We are interested in the \emph{value of the game}, i.e.\ the optimal value the players can ensure.
Formally, for an objective $\Phi : \Infinitepaths<\G> \to (\Reals \cup \infty)$, the value of state $s$ is defined as
\begin{equation*}
	\val_{\G, \Phi}(s) \eqdef {\sup}_\straa~{\inf}_\strab~\expec_{\G, s}^{\straa, \strab}[\Phi] = {\inf}_\strab~{\sup}_\straa~\expec_{\G, s}^{\straa, \strab}[\Phi],
\end{equation*}
where the latter equality holds for all objectives we consider \cite{DBLP:journals/jsyml/Martin98,DBLP:journals/ijgt/MaitraS98,DBLP:conf/memics/BrazdilKN12,DBLP:conf/lics/KieferMSW17a}, i.e.\ the games are \emph{determined}.
We consider four different objectives, namely reachability, safety, total reward, and mean payoff, which we now briefly introduce.

\paragraph{Reachability and Safety~\protect\cite{condonAlgo}} are specified by a set of target (respectively avoid) states $\targets \subseteq \allstates$.
We write $\reach \targets$ to denote all paths $\infinitepath$ that reach $\targets$, i.e.\ there exists an $i$ such that $\infinitepath_i \in \targets$.
Then, for reachability we have that $\Phi(\infinitepath) = 1$ if $\rho \in \reach \targets$ and $0$ otherwise.
Dually, in safety, the goal is to avoid the given states, i.e.\ $\Phi(\infinitepath) = 0$ if $\rho \in \reach \targets$ and $1$ otherwise.
Note that maximizing a safety objective is equivalent to minimizing a reachability objective.
We explicitly consider safety, since it allows us to illustrate some key concepts in a simpler setting.
For both, MD strategies are sufficient \cite{DBLP:conf/isaac/AnderssonM09}.

\paragraph{Total Reward~\protect\cite{SGreward}} (also called expected reward) is specified by a \emph{reward function} $\statereward : \allstates \to \RealsNonneg$.
The value of the objective is the sum of the rewards that are accumulated along the run, formally $\Phi(\infinitepath) = {\sum}_{i=0}^\infty \statereward(\infinitepath_i)$.
As rewards are non-negative, this sum always is defined, however it might be infinite.
In any case, MD strategies are sufficient \cite{SGreward}.

We remark that restricting to non-negative rewards is a standard assumption \cite{SGreward,DBLP:conf/sfm/ForejtKNP11}.
In particular, once we allow for both positive and negative rewards, several issues arise, such as negative cycles, non-converging sums, etc., see \cite[Sec.~5.2]{DBLP:books/wi/Puterman94} for further remarks.

For the sake of readability, we only consider the most common variant of total reward for now.
In \cref{sec:discussion}, we discuss the (few) modifications required to apply our methods to the other two variants defined by \cite{SGreward}. 

\paragraph{Mean payoff (or long-run average reward)~\protect\cite{gillette1957stochastic}} again is based on a reward function, but instead of the accumulated total reward, it considers the limit of the average reward.
Formally, for a reward function $\statereward : \allstates \to \RealsNonneg$ the mean payoff of a path is defined by $\Phi(\infinitepath)
\eqdef \liminf_{n \to \infty} (\frac{1}{n} \sum_{j=0}^{n-1} \statereward(\rho_j))$.
Negative rewards do not add additional complexity (see \appref{app:MP-rescale}), and we omit them for consistency.
Again, MD strategies are sufficient \cite[Thm.~1]{liggett1969stochastic}.

\paragraph{Canonical Forms}
The objectives reachability and safety do not change their value after a target (or avoid) state is reached.
Thus, without loss of generality, we assume that all states in $\targets$ are \emph{absorbing}, i.e.\ have only a single action which leads back to themselves.
For total reward, we assume that all states have finite value:
We can identify all states where the value is infinite through graph analysis \cite[Sec.~4.3]{SGreward} and remove them from the game a-priori.

\subsection{Value Iteration and Fixpoint Characterization}\label{sec:prelim-VI}
\emph{Value iteration} (VI) is a classical and versatile solution approach applied in numerous settings.
At its heart, VI relies, as the name suggests, on iteratively applying an operation to a value function, i.e.\ a mapping $f : \allstates \to \Reals$.
In the following, we call such a function \emph{assignment}, to avoid confusion with \emph{the} value of a game $\val_{\G, \Phi}$.
The shape of both the assignment and the iterated update naturally depends on the problem at hand.
Often, the update is derived from a fixpoint characterization of the assignment, which also is known as \emph{Bellman optimality equations}.
Since VI is central to this work, we briefly outline the structure of the classical value iteration solutions for our considered objectives.
We refer to \cite{DBLP:conf/spin/ChatterjeeH08} for an in-depth discussion of value iteration and provide a brief overview.

For reachability, the values satisfy the equation \cite{DBLP:journals/iandc/BrazdilBKO11}
\begin{equation}
	\val(s) = \begin{dcases*}
			1 & if $s \in \targets$, and \\
			\opt_{a \in \act(s)} {\sum}_{s' \in \allstates} \trans(s, a, s') \cdot \val(s') & otherwise,
		\end{dcases*} \label{eq:reachability_equation}
\end{equation}
Classical value iteration starts with a vector $x_0$ which assigns $1$ to all states in $\targets$ and $0$ to all others, and then iterates
\begin{equation}
	x_{i+1}(s) = \opt_{a \in \act(s)} {\sum}_{s' \in \allstates} \trans(s, a, s') \cdot x_i(s'). \label{eq:bellman}
\end{equation}
This update rule often is described as an instance of \enquote{Bellman backup} or \enquote{Bellman update}.
Iterating this rule converges to the correct value in the limit and furthermore is a monotonically improving lower bound on the true value, i.e.\ $x_i \leq x_{i+1} \leq \val_{\G, \Phi}$~\cite{DBLP:journals/iandc/BrazdilBKO11}.
Further, $x_i$ equals the optimal probability to reach the target in $i$ steps.
For the dual case, safety, we initially assign 0 to all avoid states $\fstates$ and 1 to all others and then iteratively apply \cref{eq:bellman}, obtaining a sequence converging from above.

For total reward we additionally consider state rewards.
The initial vector $x_0$ assigns $0$ everywhere and we iterate
\begin{equation}
	x_{i+1}(s) = \statereward(s) + \opt_{a \in \act(s)} {\sum}_{s' \in \allstates} \trans(s, a, s') \cdot x_i(s'). \label{eq:bellmanTR}
\end{equation}
This iteration computes the $i$-step optimal reward and converges to the true value in the limit.

For mean payoff, the classical value iteration does not compute a sequence converging to the mean payoff values.
Instead, we compute the accumulated $i$-step reward $x_i$ as in \cref{eq:bellmanTR} and then estimate the mean payoff as $x_i / i$ (or $x_i - x_{i-1}$).
This approach was shown correct for both MDP~\cite[Sec.~9.4]{DBLP:books/wi/Puterman94} and classical (non-stochastic) games~\cite[Sec.~2]{DBLP:journals/tcs/ZwickP96} and conjectured to be true for SG in~\cite[Sec.~5.2]{DBLP:conf/spin/ChatterjeeH08}.
We provide a direct proof of this conjecture
in \appref{app:mp-convergence}.
See also \appref{app:mp-useful} for more details on mean payoff in general.

We stress that there are two different notions, which however often (but not always) coincide, leading to potential confusion.
First, there are the updates performed by value iteration (\enquote{Bellman updates}), e.g.\ the right hand side of \cref{eq:bellmanTR}.
Second, there is the corresponding fixpoint characterization (\enquote{Bellman optimality equations}).
These two notions agree for, e.g., reachability, but not for mean payoff.
Both of these concepts will be relevant in this work, as well as a careful differentiation between them.
As such, we refer to the value iteration operator of an objective $\Phi$ as Bellman updates and denote it by $\textsc{VI}_{\Phi}$, whereas the right-hand side of the fixpoint characterization is referred to as fixpoint updates, written as $\fixpointup_{\Phi}$.
Thus, $x_{i+1} = \textsc{VI}_{\Phi}(x_i)$ and $\val_{\G, \Phi} = \fixpointup_{\Phi}(\val_{\G, \Phi})$.
For example, for $\Phi$ as reachability, $\textsc{VI}_{\Phi} = \fixpointup_{\Phi}$ is given by \cref{eq:bellman}; in contrast, for $\Phi$ being mean payoff, $\textsc{VI}_{\Phi}$ is given by \cref{eq:bellmanTR} while $\fixpointup_{\Phi}$ is given by \cref{eq:bellman}.


\subsection{Approximate Solutions and Bounds}
In this paper, we aim to derive an \emph{$\varepsilon$\nobreakdash-approximation} of the value.
This means that for a game $\G$, an objective $\Phi$, and a precision $\varepsilon$, we want to compute an assignment $x$ such that provably for all states $s$ we have $\abs{\val_\Phi(s) - x(s)} < \varepsilon$.
This allows us to smoothly trade precision for computation time by adapting $\varepsilon$.
Moreover, some practically efficient algorithms inherently are of approximative nature:
For many applications, VI approaches typically yield solutions converging in the limit.
However, convergence in the limit alone is not enough for our purposes, since we do not have any practical bounds on the error at any point in the iteration and thus cannot derive any $\varepsilon$-approximations.
To tackle this issue, recent works, e.g.\ \cite{atva,HM18,KKKW18}, introduced converging lower \emph{and} upper bounds.
This allows to stop the iteration once these values are sufficiently close to each other.
Already in MDP, obtaining such bounds requires non-trivial reasoning related to so-called end components, defined in the following.
One of our main contributions is to extend that reasoning, briefly summarized in the next section, to stochastic games.

\subsection{Components and Exits}
%
Intuitively, an end component is a set of states in which the system can remain forever, given suitable strategies.
Formally, a pair $(R, B)$, where $\emptyset \neq R \subseteq \allstates$ and $\emptyset \neq B \subseteq \Union_{s \in R} \act(s)$, is an \emph{end component (EC)} \cite{dA97} if
	(i)~for all $s \in R$ and $a \in \act(s) \intersection B$ we have $\post(s, a) \subseteq R$, and
	(ii)~for all $s, s' \in R$ there is a finite path $s a_0 \dots a_n s' \in (R \times B)^\star \times R$, i.e.\ the path stays inside $R$ and only uses actions in $B$.
	Inclusion-maximal ECs are called \emph{maximal end component (MEC)}.

	We say $s \in R$ is an \emph{exit} of $R$ if there is an \emph{exiting action} $a \in \act(s)$ with $\post(s, a) \nsubseteq R$, and we write $(s, a) \exits R$.
%
We identify an EC with its set of states.
Note that given two ECs $(R_1, B_1)$ and $(R_2, B_2)$ with $R_1 \intersection R_2 \neq \emptyset$, their union $(R_1 \union R_2, B_1 \union B_2)$ also is an EC.
Consequently, each state belongs to at most one MEC.
The set of MECs can be determined in polynomial time \cite{CY95}.
Moreover, independent of the strategies, the play of an MDP / SG eventually remains inside a single MEC with probability one \cite{dA97}.
We can \emph{restrict} a game $\G$ to an EC $E = (R, B)$, written $\G_{\restrict}$, by defining $\allstates' = R$, $\act' = B$, $\act'(s) = \act(s) \intersection B$ and adapting $\trans$ accordingly.
This captures all behaviours that can occur inside the EC $E$.

The corresponding notion on Markov chains are \emph{bottom strongly connected components} (BSCCs), which are sets of states $R$ that are (i)~\emph{strongly connected}, i.e.\ for every pair $s, s' \in R$ there is a non-empty finite path from $s$ to $s'$, (ii)~\emph{inclusion maximal}, i.e.\ there exists no strongly connected $R'$ with $R \subsetneq R'$, and (iii)~\emph{bottom}, i.e.\ for all $s \in C, s' \in S \setminus C$ we have $\trans(s, s') = 0$.

\section{Previous Approaches}\label{sec:example}
In this section, we illustrate the difficulties of obtaining converging bounds, as observed in recent works \cite{atva,HM18,KKKW18} and outline their key solution ideas on examples.
In the next sections, we then explain how we unify these ideas and generalize them to stochastic games. 

\subsection{Non-converging Upper Bounds in MDP}\label{sec:exampleMDP}

Recall that the iteration of \cref{eq:bellman} directly gives us correct lower bounds on the true value for any reachability objective.
Thus, our goal is to find a similar kind of iteration sequence which gives \emph{upper} bounds on the true value.
Unfortunately, simply applying \cref{eq:bellman} to upper bounds does not work.
%
%
%
%

\begin{figure}[t]
	\centering
	\begin{subfigure}{0.5\columnwidth}
		\centering
		\begin{tikzpicture}[auto]
		\node[maxstate] at (0,0) (p) {$p$};
		\node[maxstate] at (1.4,0) (q) {$q$};
		\node[maxstate] at (2.9,0.5) (s) {$s$};
		\node[maxstate] at (2.9,-0.5) (t) {$t$};
		
		\node[actionnode] at (2.25,0) (c) {};
		
		\path[directedge]
		(p) edge[bend left] node[action] {$a$} (q)
		(q) edge[bend left] node[action] {$b$} (p)
		(s) edge[loop right] (s)
		(t) edge[loop right] (t)
		;
		\path[actionedge]
		(q) edge node[action] {$c$} (c)
		;
		\path[->]
		(c) edge (s)
		(c) edge (t)
		(c) edge[out=90,in=45,looseness=1.25] (q)
		;
		\end{tikzpicture}
		\caption{The example MDP.} \label{fig:mdp_collapsing_example:mdp}
	\end{subfigure}
	\begin{subfigure}{0.45\columnwidth}
		\centering
		\begin{tikzpicture}[auto]
		\node[maxstate] at (1,0) (pq) {$\{p, q\}$};
		\node[maxstate] at (2.9,0.5) (s) {$s$};
		\node[maxstate] at (2.9,-0.5) (t) {$t$};
		
		\node[actionnode] at (2.25,0) (c) {};
		
		\path[directedge]
		(s) edge[loop right] (s)
		(t) edge[loop right] (t)
		;
		\path[actionedge]
		(pq) edge node[action] {$c$} (c)
		;
		\path[->]
		(c) edge (s)
		(c) edge (t)
		(c) edge[out=90,in=45,looseness=1.25] (pq)
		;
		\end{tikzpicture}
		\caption{The collapsed MDP.} \label{fig:mdp_collapsing_example:collapsed}
	\end{subfigure}

	\begin{subfigure}{\columnwidth}
		\caption{First few steps of value iteration applied to the example MDP.} \label{fig:mdp_collapsing_example:table}
		\centering
		\begin{tabular}{cccccc}
			$i$ & $\lb_i(p)$ & $\lb_i(q)$ & $\ub_i(p)$ & $\ub_i(q)$ & $\ub_i(\{p, q\})$ \\
			\toprule
			0 & 0 & 0 & 1 & 1 & 1 \\
			1 & 0 & $\sfrac{1}{3}$ & 1 & 1 & $\sfrac{2}{3}$ \\
			2 & $\sfrac{1}{3}$ & $\sfrac{4}{9}$ & 1 & 1 & $\sfrac{5}{9}$ \\
		\end{tabular}
	\end{subfigure}
	\caption{
		An example MDP inspired by~\protect\cite{KKKW18} to showcase the problem of obtaining upper bounds for reachability (left) and its \enquote{collapsed} counterpart (right).
	} \label{fig:mdp_collapsing_example}
\end{figure}

\begin{example} \label{ex:non_converging_mdp_bounds}
	Consider the example MDP in \cref{fig:mdp_collapsing_example:mdp} together with the reachability objective $\targets = \{t\}$.
	The iteration as given in \cref{eq:bellman} converges to the true value of $\frac{1}{2}$ in both states $p$ and $q$, as illustrated in \cref{fig:mdp_collapsing_example:table}.
	Note that it only converges in the limit, as within any finite number of steps, there is a positive chance to remain in $q$ when using action $c$.
	

	In order to obtain upper bounds, we may naively start with the greatest possible value of $\ub = 1$ everywhere. 
	However, this also is a solution to \cref{eq:reachability_equation}, i.e.\ a (non-least) fixpoint of the Bellman operator, and thus the iteration remains at this incorrect value instead of converging to the true value.
	As an additional step, we can analyse the graph and identify state $s$ as a state which cannot reach the target.
	However, even after setting its initial upper bound to $0$, the iteration does not converge:
	States $p$ and $q$ keep their upper bound of $1$ due to their cyclic dependency on each other's values. \qee
\end{example}
We informally call such a situation a \emph{spurious fixpoint}:
Our iterates converge to some value which is not the correct value of the game, due to such cyclic dependencies.
Identifying these dependencies is the key observation to obtain converging bounds.
In \cite{atva,HM18} the authors identify MECs as the root issue in MDP and propose to \enquote{collapse} them.
Their fundamental insight is that all states in a MEC have the same value for reachability.
Intuitively, once the system is in a MEC, it can ensure to eventually reach any state of this MEC with probability 1.
Hence, we can move to the exit of the MEC obtaining the best value (later defined as \emph{best exit value}) and continue from there.
In other words, MECs form an equivalence class w.r.t.\ reachability.
Collapsing essentially eliminates all internal behaviour of the MEC and replaces it with a single representative state comprising all exiting actions.
This ensures that the Bellman iteration only chooses among all exits of the MEC, removing the internal cyclic dependency.

In our example, we collapse the MEC $(\{p, q\}, \{a, b\})$ to a single state $\{p, q\}$.
The available actions of $\{p, q\}$ are those that exit the MEC, in this case only $c$, as shown in \cref{fig:mdp_collapsing_example:collapsed}.
Now, value iteration also converges from above.
Using variants of this observation, converging bounds have been obtained for all considered objectives on MDP~\cite{atva,HM18,DBLP:conf/cav/AshokCDKM17,DBLP:conf/cav/QuatmannK18}.


\subsection{Collapsing in SG with Reachability}

As outlined above, the key idea is to identify states whose values are in some sense \enquote{linked together}.
In MDP, we know that all states in a MEC have the same value, since the best exit can be reached independently of which state of the MEC the system is in.
As we can identify MECs by graph analysis, we can simply collapse them and thus remove all cyclic dependencies.
However, analysing MECs is not sufficient for SG:
The Minimizer may be able to prevent the Maximizer from reaching the best exit and vice versa.

\begin{figure}[t]
	\centering
	\captionsetup{width=.9\linewidth}
	\begin{tikzpicture}[auto]
	\node[minstate] at (0,0) (p) {$p$};
	\node[maxstate] at (-1.5,0) (q) {$q$};
	\node[maxstate] at (1.5,0) (s) {$s$};
	\node at (-2.5,0) (alpha) {$\alpha$};
	\node at (2.5,0) (beta) {$\beta$};
	
	\path[directedge]
	(p) edge[bend left] (q)
	(q) edge[bend left] (p)
	(p) edge[bend right] (s)
	(s) edge[bend right] (p)
	(q) edge (alpha)
	(s) edge (beta)
	;
	\end{tikzpicture}
	\caption{
		An example of a MEC where not all states have the same value, taken from~\protect\cite{KKKW18}.
		Circles belong to Minimizer, rectangles to Maximizer.
		For clarity, the letters $\alpha$ and $\beta$ represent the reachability values which can be obtained by taking the respective actions.
	}
	\label{fig:complCEC}
\end{figure}
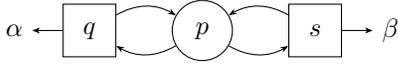

\begin{example} \label{ex:changing_secs_in_games}
	Consider the MEC in \cref{fig:complCEC}.
	First, assume that $\alpha > \beta$.
	Starting from state $s$, Maximizer cannot reach the best exit to achieve $\alpha$, since in state $p$ Minimizer can choose the action leading back to $s$.
	Thus, $p$ and $s$ have a value of $\beta$, while $q$ has the value of $\alpha$.
	If $\beta > \alpha$, we analogously get that $p$ and $q$ obtain a value of $\alpha$, while $s$ has value $\beta$.
	As such, the corresponding notion of \enquote{collapsible} states is not only graph-theoretic, but also depends on the numeric values. \qee
\end{example}

In an effort to distil the essence of MEC collapsing for stochastic games, \cite{KKKW18} introduces the notion of \emph{simple end components (SECs)}.
%
%
Intuitively, an EC $E$ is simple if all states in $E$ have the same value, i.e.\ for all $s, s' \in E$ we have that $\val_{\G, \Phi}(s) = \val_{\G, \Phi}(s')$.
As mentioned, in the MDP case each EC is simple.
Unfortunately, this is not the case in SGs, and SECs seem to be much more elusive than MECs, as we illustrate with \cref{ex:changing_secs_in_games}.
In particular, SECs do not only depend on the graph structure but also on the concrete values of each state.
However, this is what we are trying to compute in the first place!
Additionally, since we are approximating, we may \emph{never} be certain which states form a SEC.
There can be several SECs in one MEC, thus their values depend on each other and there is no topological ordering in which we could evaluate the values which we rely on (see \cref{fig:complCEC}).
Thus we cannot directly extend the idea of collapsing from MDP, as we might \enquote{commit} to a wrong guess of which states belong together.

As an alternative approach, \cite{KKKW18} propose \enquote{deflating}.
We only summarize it very briefly and provide further, related insight in the following sections, where we present our more general approaches based on this idea.
In essence, the algorithm \enquote{guesses} which states might form a SEC and then conservatively lowers the upper bound of all states in this presumed SEC.
For an intuitive view, deflating can be seen as collapsing an EC, performing an update on the \enquote{collapsed} representative, and then \enquote{un-collapsing} the EC again.
As such, we do not commit to considering a particular EC as simple forever.
It turns out that a meticulous choice of SEC-candidates is sufficient to obtain convergence.

\section{Global: Reducing to MDP} \label{sec:global}

We present our first algorithm for computing convergent under- and over-approximation of the value in SG 
in an objective-independent way.
This \enquote{global approach} elegantly delegates reasoning to existing algorithms for MDP.
The next section lifts that same idea in a more local way to the world of SG, pinpointing the root of the convergence issues.

The (surprisingly simple) key insight is as follows.
Consider a game $\G$ and an arbitrary strategy $\strab$ of Minimizer.
We fix this strategy and apply it in the SG to obtain an MDP $\G^{\cdot, \strab}$.
If $\strab$ is optimal, then playing optimally in $\G^{\cdot, \strab}$ exactly gives the value of the game.
Otherwise, if $\strab$ is sub-optimal, then Maximizer can exploit Minimizer's mistakes by answering optimally, possibly even obtaining a higher value, \emph{but certainly not lower}.
Thus, in both cases, fixing Minimizer's strategy $\strab$ and determining the value Maximizer can achieve against it, i.e.\ solving the MDP $\G^{\cdot, \strab}$, yields a \emph{correct upper bound} on the actual value.
Formally, for a strategy $\strab'$, we get
\begin{equation}
	\val_{\G, \Phi} = {\inf}_{\strab}~{\sup}_\straa~\expec_{\G, s}^{\straa, {\strab}}[\Phi] \leq {\sup}_\straa~\expec_{\G, s}^{\straa, \strab'}[\Phi] =  \val_{\G^{\cdot, \strab'}, \Phi} \label{eq:mdp_view_intuition}
\end{equation}
Dually, fixing strategies $\straa$ of Maximizer yields \emph{lower} bounds:
\begin{equation}
	\val_{\G^{\straa, \cdot}, \Phi} \leq \val_{\G, \Phi}. \label{eq:mdp_view_intuition_lower_bounds}
\end{equation}
This immediately gives rise to our first algorithm, mixing this insight together with classical VI to obtain converging bounds.

\subsection{Algorithm}\label{sec:mdp-view-algo}

\begin{algorithm}[t]
	\caption{Generic bounded value iteration based on global reduction to MDP} \label{alg:MDP-view}
	\begin{algorithmic}[1]
		\Require SG $\G$, Objective $\Phi$, initial state $\initstate$, and precision $\varepsilon$
		\Ensure $(\lb, \ub)$ such that $\lb \leq \val \leq \ub$ and $\ub(\initstate) - \lb(\initstate) \leq \varepsilon$
		\State $x_0 \gets \initViVal$\Comment{Classic VI for $\Phi$} \label{line:init} 
		\State $\lb_0(\cdot) \gets 0$, $\ub_0(\cdot) \gets \infty$ \label{line:initbounds} \Comment{Initialize bounds}
		\State $i \gets 0$
		\Repeat
		\State $i \gets i + 1$
		\medskip
		\Statex \aIndent \Commentline{Recommender procedure}
		\State $x_{i} \gets \textsc{VI}_{\Phi}(x_{i-1})$ \Comment{{Classic VI for $\Phi$}}\label{line:convergentVI}%
		\State $\straa, \strab \gets$ strategies inferred from $x_{i}$ (see \cref{lem:VIrecommend}) \label{line:infer}
		\medskip
		\Statex \aIndent \Commentline{Compute bounds (pointwise maximum / minimum)}
		\State $\lb_{i} \gets \max(\lb_{i-1}, \val(\G^{\straa,\cdot})) $ \label{line:lb}
		\State $\ub_{i} \gets \min(\ub_{i-1}, \val(\G^{\cdot,\strab})) $ \label{line:ub}
		\Until{$\ub_i(\initstate)-\lb_i(\initstate) \leq \varepsilon$} \label{line:term}
	\end{algorithmic}
\end{algorithm}

We first present the algorithm template and discuss technical details later on.
As mentioned, \cref{alg:MDP-view} is based on classic VI without bounds, initialized and performed on \cref{line:init,line:convergentVI}, respectively.
We use $\initViVal$ to denote the initialization as described in \cref{sec:prelim-VI} (e.g.\ for reachability $1$ if $s \in \targets$ and $0$ otherwise).
Additionally, we initialize and improve lower and upper bounds on the value on \cref{line:initbounds,line:lb,line:ub}, respectively, until sufficient precision is detected (\cref{line:term}).

Now, for \emph{soundness} of the algorithm, i.e.\ validity of the bounds, consider \cref{eq:mdp_view_intuition,eq:mdp_view_intuition_lower_bounds}.
Lines~\ref{line:lb}--\ref{line:ub} update the bounds this way, depending on the strategies \enquote{recommended} in \cref{line:infer}. 
No matter which strategies are recommended, the update is correct.
To ensure \emph{convergence}, it is sufficient that at some point, optimal strategies of Maximizer and Minimizer are chosen.
Indeed, if the chosen strategy $\straa$ or $\strab$ is optimal, then the values on the induced MDP equal the value of the game.
We discuss how we can efficiently derive strategies from the iterates $x_i$ in \cref{lem:VIrecommend}.


We emphasize that this observation is \emph{independent} of the objective $\Phi$ and provides a surprisingly direct way to delegate the computation of bounds for an objective to solution methods for MDP.
In a sense, identifying \enquote{problematic ECs} is done by eventually having optimal strategies inferred from $x_i$ and dealing with them is delegated to MDP reasoning. 

\subsection{Correctness and Termination}

According to the intuitive argumentation above, for the proof of correctness, we do not require \cref{line:init,line:convergentVI,line:infer} to exactly instantiate the classical value iteration.
Instead, we use the more general assumption of a \emph{recommender procedure}, which yields a sequence of strategies and eventually converges to the correct strategies.
Afterwards we show that value iteration satisfies this assumption, but also give other suggestions for the recommender procedure.
\begin{definition} \label{def:recommender}
	A \emph{(strategy) recommender} is a sequence $(\straa_i, \strab_i)$ of pairs of MD strategies such that $\straa_i$ is optimal for some $i$ and $\strab_j$ optimal for some $j$.
	A recommender is \emph{stable} if there exists an index $k$ such that $\straa_j$ and $\strab_j$ are optimal for all $j \geq k$.
\end{definition}
Note that $\straa_j$ and $\strab_j$ do not need to remain constant even for a stable recommender, they only eventually all need to be optimal.
This definition together with the previous considerations immediately yield both correctness and termination:
\begin{theorem}\label{thm:mdp-view}
	If the sequence $(\straa_i, \strab_i)$ produced on \cref{line:infer} is a strategy recommender,
	then \cref{alg:MDP-view} is correct and terminates, i.e.\ for all $i$, we have $\lb_i \leq \val_{\G, \Phi} \leq \ub_i$, and for every $\varepsilon \geq 0$, there is $i$ with $\ub_i(\initstate) - \lb_i(\initstate) \leq \varepsilon$. 
\end{theorem}
\begin{IEEEproof}
	\emph{Correctness}: Follows from \cref{eq:mdp_view_intuition,eq:mdp_view_intuition_lower_bounds}.

	\emph{Termination}: By definition of strategy recommender, the algorithm eventually evaluates the MDP induced by the optimal strategies.
	This directly gives us the true value of the game for the lower and upper bound, respectively, due to the optimality of MD strategies.
\end{IEEEproof}
Note that once optimal strategies are recommended, we obtain the exact value and may converge faster than $x_i$.
\begin{remark}
	We require a strategy recommender to yield memoryless deterministic strategies mainly for technical simplicity.
	In particular, memorylessness implies that (i)~states and end components of the induced MDP have a direct correspondence to their counterparts in the game, i.e.\ every EC in the induced MDP is also an EC in the SG, and (ii)~that only finitely many distinct recommendations exists, simplifying several arguments.
	This is not a restriction for our considered objectives, since MD strategies are sufficient for all of them and both VI as well as SI yield such strategies by default.
\end{remark}
%
%
%
%
In order to obtain a strategy recommender, we formalize a folklore result for value iteration, allowing us to obtain strategies from the iterates.
\begin{restatable}{lemma}{restatevirecommender}\label{lem:VIrecommend}
	Fix an initial pair of strategies $(\straa_0, \strab_0)$.
	For any Maximizer state $s$, define $A_i(s)$ as the set of actions in state $s$ that witness the update of the iterated vector $x_i$, i.e.\ all actions in $\argopt_{a \in \actions(s)} x_i(s, a)$.
	If $\straa_{i - 1}(s) \in A_i(s)$, define $\straa_i(s) = \straa_{i-1}(s)$.
	Otherwise, pick any $a \in A_i(s)$ and set $\straa_{i}(s) = a$.
	We analogously define $\strab_i$.
	Then, the resulting sequence $(\straa_i, \strab_i)$ is a stable strategy recommender for all considered objectives.
\end{restatable}
\begin{IEEEproof}[Proof sketch]
	Intuitively, a strategy that performs optimally for a very long time actually behaves like the infinite horizon optimal ones for our objectives.
	However, the detailed reasoning is surprisingly intricate (and objective-specific).
	In particular, keeping the previous choice if possible is important, picking any optimal action may be incorrect.
	We provide a proof together with an illustrative example in \appref{app:recommenderProof}.
	We highlight the auxiliary \cref{stm:no_stupid_ecs}, not included due to space constraints, which we believe to be useful for other works dealing with value iteration.
\end{IEEEproof}
We remark that instead of keeping the previous choice if possible, one can similarly choose a strategy randomizing uniformly over $A_i(s)$ (also discussed in \appref{app:recommenderProof}).

The desired results for our algorithm follows immediately.
\begin{theorem}
	\cref{alg:MDP-view} (as displayed with classical VI) is correct and terminates.
\end{theorem}
Note that \cref{alg:MDP-view} does not modify the intermediate computations of classical value iteration, but instead only complements them with computing lower and upper bounds (we discuss how the VI iterates can be incorporated into the bound computation for some objectives in \cref{sec:discussion}).
Solving the resulting MDPs takes polynomial time by classical encoding into linear programs~\cite{DBLP:books/wi/Puterman94}.
Consequently, any results on the speed of convergence immediately carry over from classical value iteration, i.e.\ it can take exponential time in the worst case~\cite{DBLP:conf/icalp/BalajiK0PS19} and never more~\cite{DBLP:conf/spin/ChatterjeeH08}.

\begin{remark}
	The assumptions of a recommender procedure are also satisfied by other procedures, for example, even by the naive enumeration of all strategies.
	More interestingly, it is also satisfied by classical strategy iteration, see \cite{howard,HK66,gandalf}.
	In a nutshell, SI fixes a Maximizer strategy $\straa_i$ in each iteration and computes the best response of the opponent $\strab_i$.
	Then, $\straa_{i+1}$ is obtained by picking optimal actions against $\strab_i$.
	This eventually converges to an optimal pair of strategies \cite{condonAlgo}.\footnote{
		Interestingly, improving the strategies alternatingly for both players may not converge to optimal strategies and does not yield a recommender \cite{condonAlgo}.
	}
	Thus, $(\straa_i, \strab_i)$ satisfies the conditions of a strategy recommender and \cref{alg:MDP-view} can also be used to complement strategy iteration with convergent anytime bounds.
	See \appref{app:SIversion} for pseudocode of this approach and further discussion.
\end{remark}

\section{Local: Remaining in the World of SG}

Our first algorithm addresses the convergence issues of VI by globally applying MDP reasoning, soundly under- and over-approximating the value in every iteration.
In contrast, our second algorithm, inspired by \cite{KKKW18}, avoids repeatedly solving full MDPs and instead applies specialized reasoning only locally.
This local view reveals important objective-independent concepts that allow us to pinpoint exactly where and when a problem for VI arises.
Before the technical discussion, we provide an intuitive overview and exemplify the concepts.

\begin{figure}[t]
	\centering
	\captionsetup{width=.9\linewidth}
	\begin{tikzpicture}[auto]
		\node[maxstate] at (0,0) (p) {$p$};
		\node[minstate] at (2,0) (q) {$q$};
		
		\node at (-1.5,0) (pe) {$0.1$};
		\node at (3.5,0) (qe) {$0.8$};
		
		\path[directedge]
			(p) edge (pe)
			(q) edge (qe)
			
			(p) edge[bend right] (q)
			(q) edge[bend right] (p)
		;
		\end{tikzpicture}
		\caption{
			Example fragment of an SG to illustrate fixpoint issues. 
			We omit action labels for clarity.
		} \label{fig:car}
\end{figure}
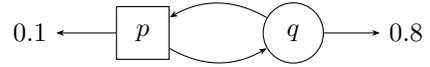

\paragraph{Why does VI not converge?} \emph{Spurious Fixpoints}.
In \cref{sec:example}, we explained that cyclic dependencies of the iterates $x_i$ prevent VI from converging, which we informally introduced as spurious fixpoints.
We discuss another example to further illustrate how these fixpoints arise and why they pose a problem for VI.
Consider the (fragment of an) SG in \cref{fig:car} with the EC $E = \{p, q\}$ and a reachability objective.
Suppose that the true values outside of the EC are as given, i.e.\ $0.1$ and $0.8$ respectively, and the iterates $x_i$ of value iteration have converged to these values outside of the EC.
The Bellman (and fixpoint) updates are $x_{i+1}(p) = \max \{0.1, x_i(q)\}$ and $x_{i+1}(q) = \min \{0.8, x_i(p)\}$.
Clearly, any value $0.1 \leq v \leq 0.8$ assigned to $p$ and $q$, i.e.\ $x_i(p) = x_i(q) = v$, yields a fixpoint thereof:
Given such $v$, both players prefer to \enquote{remain inside} the EC, relying on each others values -- a cyclic dependency -- and the iterates $x_i$ do not converge to the correct value.
Note that in particular the case of $v = 0.8$ arises for reachability upper bounds computed by VI.

\paragraph{What causes spurious fixpoints?} \emph{Simple End Components}.
Intuitively, an EC is simple if both players think that they neither gain nor lose anything by remaining in the EC for an arbitrary number of steps.
(We formally define this notion in \cref{def:simple_ec}.)
If moreover the exiting actions of the EC do not seem better to both players, they will prefer remaining inside the EC.
Concretely, in our example from \cref{fig:car}, going from state $p$ to $q$ does not reduce Maximizer's chances to reach the goal:
Minimizer can only exit and provide a higher value ($0.8$) or return back to $p$.
Dually, Minimizer is content with moving the play back to $q$ for the same reason.
Since staying for any finite number of steps is optimal, VI (applied to upper bounds) cannot determine that Maximizer actually has to leave in order to have any chance to reach the goal, as it only computes a finite-horizon probability.
Dually, for a safety objective, Minimizer does not realize that staying forever actually provides a value of 1 and leaving in $q$ is optimal.
We prove in \cref{lem:car_upgraded} that indeed SECs are the \emph{only reason} for spurious fixpoints to exist.

To further validate this intuition, assume the roles of Maximizer and Minimizer are exchanged in \cref{fig:car} and again suppose that $x_i(p) = x_i(q) = v$ for some $0.1 \leq v \leq 0.8$.
Here, no problem arises:
Minimizer immediately uses the exit of state $p$ and obtains $0.1$, while Maximizer uses the exit of $q$ to obtain the higher value of $0.8$ -- the only fixpoint is given by the true values.
Observe that the EC is not simple: Neither player wants to stay; leaving is strictly preferable for both.

\paragraph{How to handle SECs?} \emph{Staying and exit values}.
As pointed out, VI fails to handle exactly those situations where remaining inside an EC arbitrarily long is an optimal choice.
Thus, we need to devise a mechanism to inform both players about what happens by remaining inside the EC \emph{forever}.
This is formalized as \emph{staying value} (see \cref{def:staying_exiting}).
Intuitively, each player can at most achieve the staying value or the best exiting value (we formalize this intuition in \cref{lem:inflate_deflate}).
For example, in \cref{fig:car}, Maximizer cannot obtain more than the maximum of his best exit ($0.1$) and the staying value ($0$).
Thus, we can safely \emph{deflate} $x_i(p)$ to $0.1$, resolving any spurious fixpoint.
Dually, if the objective were safety and we compute lower bounds, we observe that Minimizer cannot reduce the value below the best exit ($0.8$) and the staying value ($1$), leading us to \emph{inflate} $x_i(q)$ to $0.8$.
As we show later, this treatment is both sound and sufficient to obtain convergence.

\paragraph{Summary}
We identify SECs as the central cause for non-convergence of value iteration and provide way of treating them, using staying and exit values.
The key insight is that if players are content with remaining in an EC for arbitrarily time, we need to inform them about what happens \enquote{at infinity}.


We highlight that our concepts allow us to simplify, unify, and extend findings of the last decade, from MEC collapsing \cite{atva,HM18} to deflating \cite{KKKW18} (see \cref{sec:discussion}).
Moreover, these concepts are agnostic of the objective, only requiring it to be of a very general form, which we introduce as \emph{fixpoint-linear}.

%

\subsection{Fixpoint-Linear Objectives}
To start our technical discussion, we first introduce a unified view of our considered objectives.
Recall that a central aspect (and root cause for spurious fixpoints) is the fixpoint characterization of each objective.
By considering the shape thereof, we classify a broad spectrum of \emph{fixpoint-linear} objectives.

Each of the discussed objectives $\Phi$ has an associated \emph{fixpoint characterization}, i.e.\ a function $\fixpointup_\Phi$ which given an assignment $f$ and state $s$ yields the supposed value for this state based on its successors.
Concretely, we have $\fixpointup_\Phi(f, s) \eqdef \opt_{a \in \act(s)} f(s, a)$ for reachability, safety, and mean payoff\footnote{
	Recall that classical value iteration for mean payoff uses a different update, as explained in \cref{sec:prelim-VI}.
}, and $\fixpointup_\Phi(f, s) \eqdef \opt_{a \in \act(s)} \statereward(s) + f(s, a)$ for total reward.
Note that these are all of quite similar, linear shape.
%
\begin{definition} \label{def:fixpoint_linear}
	An objective $\Phi$ is called fixpoint-linear if for every game $\G$ (in canonical form), the value is a solution to an (affine) linear equality of the form
	\begin{align*}
		f(s) & = \opt_{a \in \act(s)} \fpoff_\Phi(s) + f(s, a)                                                \\
		     & = \opt_{a \in \act(s)} \fpoff_\Phi(s) + {\sum}_{s' \in \allstates} \trans(s, a, s') \cdot f(s')
	\end{align*}
	for some $\fpoff_\Phi(s) \geq 0$ in every state $s$.\footnote{We deliberately exclude discounting objectives in this definition due to technical reasons and discuss how they can be handled in \cref{sec:discussion}.}
	We call the right hand side \emph{fixpoint update of $\Phi$}, written $\fixpointup_\Phi(f, s)$ for a function $f$. 
\end{definition}
Note that $\fpoff_\Phi(s) = \statereward(s)$ for total reward and $\fpoff_\Phi \equiv 0$ for all others.
Many classical objectives are of this form, and even more can be reduced to this; see \cref{sec:discussion} for discussion.
However, this certainly is not true for all objectives, such as the (non-linear) conditional value-at-risk \cite{DBLP:conf/lics/KretinskyM18,DBLP:conf/aaai/Meggendorfer22}.

%


\subsection{Which End Components are Simple?}
With these notions at hand, we introduce our generalization of simple end components.
Throughout this section, we fix a game $\G$ and fixpoint-linear objective $\Phi$.

Our intuitive overview identifies two conditions which jointly allow spurious fixpoints to arise: we may run into problems if (i)~inside an EC an assignment $f$ is a fixpoint and (ii)~the actions of the EC are preferred over exiting actions by both players.
As we confirm later, $f$ might be a spurious fixpoint exactly in this case.
We begin by showing that for fixpoint-linear objectives, end components which exhibit such a \enquote{fixpoint behaviour} already have a very specific shape:
\begin{restatable}{lemma}{restatesecswithoutreward} \label{stm:secs_are_without_reward}
	Let $f$ an assignment and $E = (R, B)$ an EC.
	Then
	\begin{equation*}
		f(s) = \fpoff_\Phi(s) + f(s, a) \text{ for all $s \in R, a \in \actions(s) \intersection B$},
	\end{equation*}
	implies $\fpoff_\Phi(s) = 0$ for all $s \in R$.
\end{restatable}
See \appref{app:proof_secs_without_reward} for a formal proof.

The first condition means that $f$ satisfies the fixpoint equation for $\Phi$ on $\G_{\restrict}$, i.e.\ the game restricted to $E$.
With our intuition in mind, this means that only ECs where $\fpoff_\Phi(s) = 0$ could be a problem.
This motivates the following definition.
\begin{definition} \label{def:simple_ec}
	Let $E = (R, B)$ an EC and $f$ an assignment.
	We call $E$ a \emph{simple end component for $f$} (SEC for $f$) if
	\begin{equation*}
		f(s) = f(s, a) \text{ for all $s \in R, a \in \actions(s) \intersection B$}.
	\end{equation*}
	Inclusion maximal SECs are called \emph{maximal simple end component} (MSEC).
	We write \enquote{SEC} instead of \enquote{SEC \emph{for $\val_{\G, \Phi}$}}.
\end{definition}
In \cite{KKKW18}, SECs are introduced as ECs where all states have the same value, which for reachability implied that the EC potentially allows for spurious fixpoints.
It turns out that our definition of SECs implies that view.
\begin{restatable}{corollary}{restatesecsequality} \label{stm:secs_assignment_equal}
	Let $f$ an assignment and $E$ an EC.
	Then, $E$ is a SEC for $f$ if and only if (i)~$f(s) = f(s')$ and (ii)~$\fpoff_\Phi(s) = 0$ for all $s, s' \in E$.
\end{restatable}
See \appref{app:proof_assignment_equal} for a formal proof.
This motivates naming such ECs \enquote{simple}:
All states have the same value and no rewards are gained; arguably as simple as an EC can be.

Matching the discussed intuition, SECs indeed exactly pinpoint spurious fixpoints:
\begin{restatable}{lemma}{restatecar} \label{lem:car_upgraded}
	Let $f$ a fixpoint for $\fixpointup_\Phi$.
	If $f \not\equiv \val_{\G, \Phi}$, then there exists a SEC $E$ for $f$ where $f(s) \neq \val_{\G, \Phi}(s)$ for all $s \in E$.
\end{restatable}
The proof can be found in \appref{app:proof_car}.


Recall that spurious fixpoints are the sole reason why fixpoint updates do not converge.
Additionally, \cref{lem:car_upgraded} shows that the only reason why a spurious fixpoint $f$ can exist is due to SECs:
If $f$ is fixpoint for $\fixpointup_\Phi$, i.e.\ applying fixpoint updates does not change $f$, then $f \equiv \val$ if and only if $f(s) = \val(s)$ on all SECs for $f$.
Thus, we identified the root cause for spurious fixpoints.
However, we do not yet have a tool to deal with them.
As discussed, SECs are exactly those places where both players think they can delay the play indefinitely and we need to inform them what happens if they indeed stay forever.

\subsection{Should I Stay or Should I Go?}
%

In order to distinguish the case of (eventually) leaving and staying forever, we introduce the notions of staying value and best exit, and treat them through \emph{deflating} and \emph{inflating}.
In a sense, these correspond to offering a glimpse at what could happen in the infinite horizon.
\begin{definition} \label{def:staying_exiting}
	Fix an EC $E = (R, B)$.
	The staying value of a state $s \in E$ is defined as
	\begin{equation*}
		\stayVal_{\G, \Phi}(E,s) \eqdef \val_{\G_{\restrict}, \Phi}(s),
	\end{equation*}
	where $\G_{\restrict}$ is the SG $\G$ restricted to states and actions in $E$.

	The \emph{best exit value} (from $E$ under an assignment $f$) of Maximizer and Minimizer are defined as
	\begin{align*}
		\bE^\Box_f(E) & \eqdef {\max}_{s \in R \intersection \allstates_\Box, (s,a) \exits E} f(s,a) \quad \text{and} \\
		\bE^\circ_f(E) & \eqdef {\min}_{s \in R \intersection \allstates_\circ, (s,a) \exits E} f(s,a).
	\end{align*}
\end{definition}
Note that we defined $\max \emptyset = - \infty$ and $\min \emptyset = \infty$, i.e.\ the \enquote{best exit} of Maximizer in an EC without exits is $-\infty$.
\begin{figure}[t]
	\centering
	\captionsetup{width=.9\linewidth}
	\begin{tikzpicture}[auto]
		\node[maxstate] at (0,0) (p) {$p \mid 0$};
		\node[maxstate] at (3,0.5) (q) {$q \mid 3$};
		\node[minstate] at (1.75,-1) (s) {$s \mid 2$};
		
		\node at (4.5,0.5) (qe) {$6$};
		\node at (3.25,-1) (se) {$0$};
		\node at (-1.5,0) (pe) {$0$};
		
		\path[directedge]
			(p) edge (pe)
			(q) edge (qe)
			(s) edge (se)
			
			(p) edge[bend right] (s)
			(s) edge[bend right] (p)
			(s) edge (q)
			(q) edge[loop above] (q)
			(q) edge[bend right] (p)
		;
	\end{tikzpicture}
	\caption{
		Example stochastic game to illustrate the definition of staying and exit value.
		We write the reward assigned to each state next to its name.
	} \label{fig:stay_and_exit}
\end{figure}
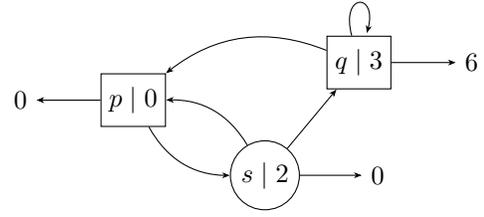
\begin{example} \label{ex:stay_and_exit}
	Consider the EC $E = \{p, q, s\}$ of the SG from \cref{fig:stay_and_exit} together with a mean payoff objective.
	States $p$ and $s$ have a staying value of $1$ in $E$:
	When restricted to the MEC, Minimizer always moves back to $p$ instead of going to $q$.
	State $q$ has a staying value of $3$ by virtue of its self-loop.
	The best exit of Maximizer is given by $6$ in state $q$, which is also what Maximizer can realize in $q$.
	State $p$ has a value of $0$ (which is strictly less than both staying value and Maximizer's best exit), since the Minimizer chooses to exit the EC in state $s$, using Minimizer's best exit. \qee
\end{example}
Recall our intuition that issues can arise exactly when both players prefer staying over leaving under the current values.
Given our definitions, we can now formalize this claim.
\begin{lemma} \label{lem:car_spoiler}
	Let $f$ an assignment and $E$ a SEC for $f$.
	We have $\bE^\Box_f(E) \leq f(s) \leq \bE^\circ_f(E)$ for any $s \in E$ \emph{if and only if} a fixpoint update does not modify the value of $f$ on the EC.
\end{lemma}
\begin{IEEEproof}
	Since $E$ is SEC, by \cref{stm:secs_are_without_reward}, we have that $f(s) = f(s') = c$ for all $s \in E$.

	Observe that the backward direction follows immediately:
	If $f$ is not modified on the EC it means that in every state the value $c$ was weakly preferred over all exiting options.

	For the forward direction, fix any Maximizer state $s$.
	Since $\bE^\Box_f(E) \leq c$, we also have that $f(s, a) \leq f(s, b)$ for any $(s, a)$ exiting $E$ and $(s, b)$ remaining in $E$.
	This dually holds for Minimizer states.
	Consequently, the fixpoint update always considers the interior actions, yielding $\fixpointup_\Phi(f, s) = c = f(s)$ for every $s \in E$.
\end{IEEEproof}
We apply this lemma to our running example.
\begin{example} \label{ex:car}
	Recall the example of \cref{fig:car}.
	Then, we have $\bE^\Box_{x_i}(E) = 0.1 < 0.8 = \bE^\circ_{x_i}(E)$.
	As discussed, any value $0.1 \leq v \leq 0.8$ assigned to $p$ and $q$ yields a fixpoint of $\fixpointup_\Phi$ (which is exactly what \cref{lem:car_spoiler} yields).
	\qee
\end{example}
So, if both players believe they prefer staying based on the current assignment, we may run into trouble.
To resolve this, we want to inform them about the consequences of staying.
We show that in a SEC, the value of any state is bounded from above by the maximum of its staying value and the best exit value of Maximizer.
Naturally, the dual statement holds for the Minimizer.
\begin{restatable}{lemma}{restateinflatedeflate} \label{lem:inflate_deflate}
	Fix a memoryless Minimizer strategy $\strab$.
	Let $E = (R, B)$ be a SEC and $f$ an upper bound on the value $\val_{\G, \Phi}$.
	Then, we have for every $s \in R$
	\begin{equation}
		\val_{\G, \Phi}(s) \leq \max(\stayVal_{\G^{\cdot, \strab}, \Phi}(E, s), \bE^\Box_f(E)). \label{eq:defl}
	\end{equation}
	Dually, for a Maximizer strategy $\straa$ and $f$ lower bound, we get
	\begin{equation}
		\val_{\G, \Phi}(s) \geq \min(\stayVal_{\G^{\straa, \cdot}, \Phi}(E, s), \bE^\circ_f(E)). \label{eq:infl}
	\end{equation}
\end{restatable}
The (rather technical but objective-independent) proof can be found in \appref{app:proof_inflate_deflate}.
Note that the actual value may even be strictly smaller than both staying value and best exit, for example when Minimizer can prevent reaching the best exit, see \cref{ex:stay_and_exit}.
We call applying \cref{eq:defl} \emph{deflating} (reducing potentially \enquote{blown up} values) and dually \cref{eq:infl} \emph{inflating}.

With this idea at hand, we derive further insights for our considered objectives.
\begin{remark} \label{rem:reachability_deflate}
	For reachability, observe that the staying value in all non-target SECs is $0$ and iterates provide a converging lower bound.
	Thus, \cref{eq:defl} yields $\val_{\G, \Phi}(s) \leq \bE^\Box_{x_i}(E)$:
	The best value Maximizer can achieve is at most the best value achievable by leaving the EC.
	To visualize this further, consider \cref{fig:car} again.
	We see that $\bE^\Box_{x_i}(E) = 0.1$ which indeed is an upper bound to the correct value:
	Maximizer may think taking the staying action is not harmful -- either Minimizer \enquote{gives} $0.8$ or the play returns back to state $p$.
	However, by revealing the consequences of doing this forever (the actual staying value), Maximizer realizes that instead the best exit is the best possible chance in this EC, reducing the upper bound to $0.1$.
	Moreover, applying this update in state $p$ is enough to resolve convergence problems.
	This holds in general:
	Since for reachability iterates converge to the correct value from below on their own, we never need to inflate them.

	In the case of safety, we dually observe that the staying value is $1$ and iterates converge from above, meaning that only inflation is necessary.
	This contrasts mean payoff, where the staying value could actually lie between the best exits and we need to consider both sides.
\end{remark}
%
%
%
To summarize, \cref{lem:car_spoiler} identifies SEC as potential problems, while \cref{lem:inflate_deflate} provides us with a sound, conservative way of treating these \enquote{alarms}.
In the following, we combine this insight with classical value iteration, which gives us \enquote{completeness} in the form of termination guarantees.
We emphasize that up until now, our reasoning holds for all fixpoint-linear objectives.

\subsection{Algorithm}\label{sec:local-algo}

\begin{algorithm}[t]
	\caption{Bounded value iteration with local reasoning}\label{alg:SG-view}
	\begin{algorithmic}[1]
		\Require SG $\G$, Objective $\Phi$, initial state $\initstate$, and precision $\varepsilon$
		\Ensure $(\lb, \ub)$ such that $\lb \leq \val \leq \ub$ and $\ub(\initstate) - \lb(\initstate) \leq \varepsilon$
		\State $x_0 \gets \initViVal$ \Comment{Classic VI for $\Phi$} \label{line2:init}
		\State $\lb_0(\cdot) \gets 0$, $\ub_0(\cdot) \gets \upperViVal$ \label{line2:initbounds} \Comment{Initialize bounds}
		\State $i \gets 0$
		\While{$\ub_i(\initstate) - \lb_i(\initstate) > \varepsilon$}\label{line2:term}
			\State $i \gets i + 1$ 
			\medskip
			
			\Statex \aIndent \Commentline{Convergent recommender procedure}
			\State $x_i \gets \text{VI}_\Phi(x_{i-1})$ \Comment{Classic converging VI for $\Phi$} \label{line2:convergentVI}
			\State $\straa, \strab \gets \text{ strategies inferred from } x_i$ (see \cref{lem:VIrecommend}) \label{line2:infer}
			\medskip
			
			\Statex \aIndent \Commentline{Fixpoint updates for $\Phi$ on \emph{bounds} in \emph{induced MDPs}}
			\State $\hat{\lb}_i \gets \min(\lb_i, \fixpointup_\Phi^{\G^{\straa, \cdot}}(\lb_{i-1}))$ \label{line2:BU-L}
			\State $\hat{\ub}_i \gets \max(\ub_i, \fixpointup_\Phi^{\G^{\cdot, \strab}}(\ub_{i-1}))$ \label{line2:BU-U}
			\medskip
			
			\Statex \aIndent \Commentline{Additional local MDP reasoning, de- \& inflate}
			\For {each MSEC $E$ in $\G^{\cdot, \strab}$} \label{line2:defl-for}
				\For {all $s \in E$}
					\State $\mathcal{U}_i(s) \gets \max(\stayVal_{\G^{\cdot, \strab},\Phi}(E,s), \bE^\Box_{\hat{\ub}_i}(E))$ \label{line2:defl-UM}
					\State $\ub_i(s) \gets \min(\hat{\ub}_{i}(s), \mathcal{U}_i(s))$ \label{line2:defl-upd}
				\EndFor
			\EndFor
			\For {each MSEC $E$ in $\G^{\straa, \cdot}$} \label{line2:infl-for}
				\For {all $s \in E$}
					\State $\mathcal{L}_i(s) \gets \min(\stayVal_{\G^{\straa, \cdot},\Phi}(E,s), \bE^\circ_{\hat{\lb}_i}(E))$ \label{line2:infl-LM}
					\State $\lb_i(s) \gets \max(\hat{\lb}_{i}(s), \mathcal{L}_i(s))$ \label{line2:infl-upd}
				\EndFor
			\EndFor
			\State Set $\ub_i(s) \gets \hat{\ub}_i(s)$ and $\lb_i(s) \gets \hat{\lb}_i(s)$ if not defined
		\EndWhile
	\end{algorithmic}
\end{algorithm}

With these intuitions in mind, we now present our \enquote{local} solution approach in \cref{alg:SG-view}.
As in \cref{alg:MDP-view}, \cref{line2:init,line2:convergentVI,line2:infer} describe the classic convergent VI procedure that is used to recommend strategies.
However, the bounds $\lb_i$ and $\ub_i$ are not computed by solving an MDP.
Instead, in \cref{line2:BU-L,line2:BU-U} we perform fixpoint updates on the bounds, i.e.\ \cref{eq:bellmanTR} for total reward and \cref{eq:bellman} otherwise, to obtain $\hat{\lb}_i$ and $\hat{\ub}_i$.
Since these might not converge, as shown in \cref{ex:non_converging_mdp_bounds}, we apply the additional local reasoning in the loops of \cref{line2:defl-UM,line2:infl-LM}, obtaining $\mathcal{L}_i$ and $\mathcal{U}_i$, respectively.

We emphasize that this algorithm conceptually handles all the different objectives we consider.
Independent of objective, the computation repeatedly (i)~applies fixpoint updates and (ii)~adjusts the approximations in problematic ECs so they are dependent on exit and staying values.
Observe that we search for SECs in the MDP induced by the recommended strategies.
This corresponds to the \enquote{guess} for SECs, as in \cite{KKKW18}.
Since the updates we perform are conservative, guessing wrongly does not hurt by the reasoning of the previous section, in particular \cref{eq:mdp_view_intuition,eq:mdp_view_intuition_lower_bounds}.
It remains to discuss three sub-procedures.

\paragraph*{Initializing Upper Bounds ($\upperViVal$)}
Firstly, we cannot initialize the upper bounds to $\infty$ everywhere, as this yields a fixpoint.
Fortunately, for reachability, safety, and mean payoff, we can directly derive an upper bound ($1$ or the maximal occurring reward, respectively).
For total reward, we can compute a finite upper bound on the total reward for all states as in~\cite{SGreward}.
See \appref{app:init-vi} for details.

\paragraph*{Finding MSECs}
We only need to find MSECs in the induced MDPs, which is straightforward in almost all cases:
For reachability, safety, and mean payoff, those are exactly the MECs of the MDP by virtue of \cref{stm:secs_are_without_reward}.
For total reward, we need to employ further care:
We excluded that states have a value of infinity \emph{in the game}, however, Maximizer could obtain infinite reward in $\G^{\cdot, \strab}$ due to a sub-optimal Minimizer strategy.
However, any MEC in the MDP either has value 0 or infinity, thus, still every MEC is MSEC in the MDP.
For the Minimizer, observe that no state can have a value of infinity in $\G^{\straa, \cdot}$ by our assumption:
Against a sub-optimal strategy of Maximizer, Minimizer can achieve even smaller values.
Thus, MSECs are all maximal end components that remain after removing all states with non-zero rewards, again by \cref{stm:secs_are_without_reward}.

\paragraph*{Computing the Staying Value}
For all objectives except mean payoff, simple graph analysis is sufficient, see \appref{app:MP-stayVal}.
For mean payoff, an exact computation is possible, however we additionally discuss how we can obtain convergent under- and over-approximations from the iterates $x_i$.
In particular, we can interleave the computation of staying values to finer and finer precisions with the propagation of bounds, similar to the \emph{on-demand} adaptation of VI of \cite{DBLP:conf/cav/AshokCDKM17}.
When we replace the exact value with converging upper and lower bounds, the correctness and convergence guarantees do not change.
We describe the details of both the exact as well as the approximate computation in \appref{app:MP-stayVal}.



Now that the algorithm was intuitively explained, we formally state its effectiveness.
The technical proof can be found in \appref{app:proof_sg_correct}.
\begin{restatable}{theorem}{restatesgcorrect} \label{stm:algo_sg_correct}
	\Cref{alg:SG-view} is correct and terminates for any considered objective $\Phi$.
\end{restatable}
Note that \cref{alg:SG-view} again only relies on a (stable) recommender procedure and is not bound to value iteration.

%
%


%

\section{Discussion}\label{sec:discussion}

To conclude, we provide several remarks on our algorithms for further insights and relate them to previous approaches.


\subsection{Monotonicity}
In 
both algorithms, we update bounds by additionally comparing with the previous lower respectively upper estimate to ensure monotonicity.
This property is preferable, because a user of the algorithm naturally expects convergent anytime bounds to be monotonically improving.
However, for the correctness and convergence this comparison is not necessary if the strategy recommender is stable.


\subsection{Relation between \cref{alg:MDP-view,alg:SG-view}}
Both algorithms rely on a recommender procedure that eventually suggests optimal strategies, and both of them use the suggested strategies to apply specialized reasoning to deal with spurious fixpoints in ECs.
However, \cref{alg:MDP-view} globally applies this reasoning by completely solving both induced MDPs every iteration.
This allows us to transparently re-use existing methods for MDP solving.
In contrast, \cref{alg:SG-view} performs the fixpoint updates on the bounds and then only applies the specialized reasoning locally to resolve cyclic dependencies.
This way, \cref{alg:SG-view} avoids solving large parts of the MDP repeatedly which do not need specialized reasoning.
However, it cannot immediately obtain the values when an optimal strategy is recommended, and might only converge in the limit.

\subsection{Relationship to classical VI}
For reachability, recall that the sequence $(x_i)_{i\in\Naturals}$ already is a convergent lower bound on the value.
Consequently, separate computation of lower bounds 
can be omitted, including considering the induced MDP. 
More generally, we can directly use $x_i$ in place of $\lb_i$.
The same holds for total reward~\cite{SGreward}.
Dually, for safety 
the upper bound computation can be omitted.
Finally, for mean payoff, the values $x_i / i$ of the classic VI can oscillate around the actual value, hence both bounds have to be computed in addition to $x_i$.


\subsection{Relationship to \cite{KKKW18}}
In essence, \cite{KKKW18} performs a structural analysis on MECs using the current approximations and then \emph{deflates}, i.e.\ decreases the upper bounds to values of some Maximizer's (MEC-exiting) actions.
From our perspective, it applies Minimizer's strategy $\strab$ and computes the best exits in the induced MDP.
In the simpler case of reachability, the staying value is $0$ in all non-trivial ECs and each MEC is also MSEC.
Combining with the above insights, we obtain the algorithm of~\cite{KKKW18} as special case of \cref{alg:SG-view}.

\subsection{Relationship to \cite{widestPaths}}
The convergent anytime algorithm for reachability in SG from~\cite{KKKW18} has been extended to improve performance in~\cite{widestPaths}.
In that work, the authors compute the over-approximation by constructing the MDP $\G^{\cdot,\strab}$ as in \cref{alg:MDP-view}.
However, instead of solving the MDP completely, they over-approximate its value by turning it into a weighted graph and computing a widest path problem.
The resulting values over-approximate the value in the MDP, and hence the values in the SG.
Eventually the weights in the graph are precise enough that the algorithm converges.
Thus, to make \cref{alg:MDP-view} practically more efficient, one can replace the complete MDP-solving also with the weighted-graph-based approximation from~\cite{widestPaths}.

\subsection{Relationship to EC collapsing}
The presented insights, in particular \cref{alg:SG-view}, capture and unify all previous treatment of problematic fixpoints in MDP.
Indeed, \cref{lem:inflate_deflate} also holds for MDP.
For the objectives reachability, safety, and mean payoff, \cref{eq:defl,eq:infl} actually yield an equality, as there is no second player that can interfere.
Moreover, for these objectives on MDP, every EC is simple.
In other words, \cref{lem:inflate_deflate} yields that the value of a (simple) EC exactly equals the maximum of staying and leaving value.
Previous works~\cite{DBLP:conf/cav/AshokCDKM17} dealt with such objectives on MDP by collapsing ECs, i.e.\ replacing them with a single state that has all outgoing actions available and, in the case of mean payoff, a special \enquote{stay} action, which represents the value that can be obtained by remaining inside the EC.
Observe the exact match with the above equations:
By keeping all exiting actions, the best exit remains available to the player (at the time the EC is collapsed, the precise values of actions are not known, hence all exits need to be kept).
Analogously, by adding the stay action, the staying value may still be obtained.
Thus, applying the fixpoint update on the collapsed MDP takes the maximum over staying and all possible exits.


%


\subsection{Discounted and Bounded Objectives}
A prominent class of objectives are \emph{discounting objectives}.
For example, for discounted total reward, the value of a path is given by $\Phi(\infinitepath) = {\sum}_{i=0}^\infty \gamma^i \statereward(\infinitepath_i)$ for some discounting factor $\gamma < 1$.
While our methods as presented are not immediately applicable -- these objectives are not fixpoint-linear -- there is a classical reduction for many discounted objectives to their non-discounted counterpart.
For example, with total reward we (i)~add a trap state $\underline{s}$ to the game with $\statereward(\underline{s}) = 0$, (ii)~rescale every transition by $(1 - \gamma)$, and (iii)~add a $\gamma$-probability transition to $\underline{s}$.
By applying this technique, \cref{alg:SG-view} also is applicable to \enquote{discounting fixpoint-linear objectives}, where $\fixpointup_\Phi(f, s) = \opt_{a \in \act(s)} \fpoff_\Phi(s) + \gamma(s) \cdot \fixpointup_\Phi(f, s, a)$ with some (potentially state-dependent) discounting factor $\gamma(s) \in [0, 1]$.
(Note that this requires a sensible reward value for $\underline{s}$.)

Similarly, \emph{bounded objectives}, requiring, for example, that a goal state is reached before a time- or resource-bound is exceeded, can be handled by encoding the current usage into the state space.
For step-/time-bounded objectives, no non-trivial ECs exist and classical VI already provides a complete solution:
Iterating the Bellman operator $K$ times yields the optimal value after $K$ steps.
However, for e.g.\ cost-bounded reachability there might be ECs comprising states with cost zero, potentially inducing a SEC that needs to be dealt with.

We highlight that both types of objectives are known to have a unique fixpoint and the problem of non-converging bounds does not arise.
Yet, our framework provides a simple, objective independent proof for a wide class of objectives.

\subsection{Other Variants of Total Reward}
We now comment on adaptations to other variants of total reward.
Aside from the \enquote{classical} total reward, where rewards are simply accumulated along the path, \cite{SGreward} introduces two other variants, which we refer to as $\mathsf{TR}_0$ and $\mathsf{TR}_\infty$.
Both are defined relative to a target set $\targets$, and $\mathsf{TR}_0(\rho) = \mathsf{TR}_\infty(\rho) = \sum_{i=1}^k \statereward(\rho_k)$ if $\rho \in \reach \targets$ and $k$ is the first time the target set is reached; otherwise $\mathsf{TR}_0(\rho) = 0$ and $\mathsf{TR}_\infty(\rho) = \infty$.
Informally, we only obtain the accumulated reward if the target is reached, otherwise we get $0$ or $\infty$, respectively.
While $\mathsf{TR}_0$ can be directly reduced to our variant \cite[Sec.~4.3.3]{SGreward}, $\mathsf{TR}_\infty(\rho)$ provides an interesting, \enquote{dual} variant.
In particular, while for regular total reward it is Maximizer who wants to leave end components eventually, here Minimizer wants to leave, since otherwise $\infty$ is obtained.

As with the other objectives, we can again assume a canonical form.
In particular, we require that no states have a value of infinity -- exactly those from which Maximizer can ensure that the target states are not reached.
We remark that nevertheless this objective is different from regular total reward, since Maximizer that can force Minimizer to take \enquote{large} exits, as also explained below.

We now discuss the simple adaptations required to show that our algorithms apply to $\mathsf{TR}_\infty$, too, in form of a \enquote{checklist}.
\begin{itemize}
	\item[\emph{Recommender}:]
	It is known that VI monotonically converges to the correct value from above \cite{SGreward}.
	To obtain \cref{lem:VIrecommend}, i.e.\ that value iteration recommends correct strategies, we can proceed completely analogous to the regular total reward proof (see \appref{app:recommenderProof}).
	In particular, by convergence guarantees, we automatically get that strategies are safe eventually.
\end{itemize}
This is already enough to show correctness and convergence for \cref{alg:MDP-view}!
For \cref{alg:SG-view}, only minor points remain -- most of the reasoning in \cref{sec:local-algo} is objective independent.
\begin{itemize}
	\item[\emph{Fixpoint-Linear}:]
	This clearly holds for this objective.

	\item[\emph{Init and MSECs}:]
	Both initialization of value iteration (see \appref{app:init-vi}) and identification of MSECs is analogous to regular total reward.
\end{itemize}
%
%
As an interesting observation, since $\mathsf{TR}_\infty$ is fixpoint-linear, note that \cref{lem:inflate_deflate} holds.
Since the staying value is $\infty$ (compared to $0$ for regular total reward) and value iteration converges from above, we now only need to inflate instead of deflate, i.e.\ adjust Minimizers values inside SECs instead of Maximizer.
Intuitively, Maximizer has the \enquote{momentum} in ECs and Minimizer is forced to eventually make a move.
This is analogous to \cref{rem:reachability_deflate}, $\mathsf{TR}_\infty$ is like safety in this regard: a greatest fixpoint objective requiring only inflation.
In contrast, reachability and \enquote{classical} total reward are least fixed point objectives and thus require only deflation.

\section{Conclusion}



We have provided stopping criteria for SG with frequently used quantitative objectives.
Our approach comes in two flavours.
The first, generic algorithm allows to lift MDP-based reasoning to SG on a broad variety of objectives, requiring only very mild assumptions.
The second, local-reasoning algorithm provides a unifying solution to previously as well as newly studied objectives, and, by encompassing previous solutions and tweaks, unifies a decade of advances on MDP and SG.
More generally, we provide a solid theoretical foundation for obtaining converging bounds through value iteration for infinite-horizon objectives on stochastic games.
We conjecture that it can be directly extended to $\omega$-regular objectives (with the staying value of parity SG being computationally more complicated) or to multiple objectives (similar to multi-dimensional reachability \cite{DBLP:conf/lics/AshokCKWW20}).
Combining with other recent advances such as guessing \cite{DBLP:conf/soda/ChatterjeeMSS23} may also provide further insights (e.g.\ guessing the value of an entire SEC).

Since this work can serve as a basis for guaranteed and yet practically fast algorithms, it is desirable to extend the algorithms with adaptations of heuristics in order to foster the practical applicability.
On the one hand, topological VI \cite{DBLP:journals/jair/DaiMWG11} or better initial values \cite{gandalf} are directly applicable.
On the other hand, extending sound VI \cite{DBLP:conf/cav/QuatmannK18} or optimistic VI \cite{DBLP:conf/cav/HartmannsK20} to SG or using the widest path approach \cite{widestPaths} seem promising future directions within reach.
Besides, converging bounds allow for a sensible use of asynchronous value iteration and partial exploration, e.g.\ \cite{brtdp,atva,KKKW18}, which can then be guided by unreliable but potentially powerful learning heuristics.
Once these improvements are properly put in place, their implementation and experimental comparison will be an interesting next step towards practically scalable solutions.

We conjecture that our reasoning can be directly extended to objectives requiring finite memory.
Moreover, for many objectives requiring memory, e.g.\ $\omega$-regular specifications, the required memory structure is known a-priori and the problem can be converted to a \enquote{memoryless} one by adding the memory to the state space of the game.

%

\bibliographystyle{IEEEtran}
\bibliography{ref}

\cleardoublepage
\appendices

%

\section{Properties of Mean Payoff}\label{app:mp-useful}

We first give some intuition about the mean payoff objective.
There are two fundamental notions, as described in e.g.\ \cite{DBLP:books/wi/Puterman94}, named \emph{gain} and \emph{bias}.
Recall that each run of the system induces a sequence of rewards for mean payoff, say $10\ 10\ 5\ 10\ 5\ 3\ 5\ 3\ 5\ 3\ 5\ \cdots$.
The gain of this run is $4$ since it repeats $3$ and $5$ infinitely.
However, in the beginning, before it \enquote{stabilizes} to this average of $4$, the run achieves larger values, in particular several $10$s.
The bias of this run now equals the \emph{total} deviation its gain, i.e.\ how its \enquote{transient} or \enquote{finite} behaviour differs from the infinite part (the gain).
Note that almost all runs eventually end up in a MEC where then the gain is obtained.
An alternative view to visualize the interplay of gain and bias is that, in the limit, the $n$-step total reward of a run equals $n$ times the gain plus the bias, i.e.\ $v_n - n \cdot g + h = \mathcal{O}(1)$, where $v_n$ denotes the $n$-step total reward of a run, $g$ the gain and $h$ the bias.
The $\mathcal{O}(1)$ term (opposed to constant $0$) stems from potential small fluctuations which we do not discuss in detail now.
This view actually motivates the two approaches to compute mean payoff.
Both considering the differences in increments as well as $v_n / n$ yields the gain (where the former needs simple treatment of the $\mathcal{O}(1)$ term).

For Markov chains, we observe that BSCCs exactly mark the point where runs switch to the \enquote{infinite} regime, i.e.\ start \enquote{obtaining} their gain.
On MDP, MECs generalize this idea:
All states in a MEC have the same value since we can always move to any potential state from which an optimal value can be obtained with probability 1 and then replicate the strategy obtaining this value.
Using this realization, \cite{DBLP:conf/cav/AshokCDKM17} decompose the MDP into MECs, compute the optimal value achievable in each MEC (assuming the system is restricted to it), and then solve a \enquote{weighted reachability} query.
Essentially, they treat each MEC as a soft target where the system can decide to \emph{stay} and obtain the MECs value (replaying the optimal strategy) or move on, trying to get to a better MEC.
Since under any strategy we always end up in MECs with probability 1, MECs are the only place where the system really \enquote{obtains} the mean payoff and anything prior essentially is a matter of reachability.

In the following, we lift some basic properties of mean payoff and several of these ideas to the world of SGs, with the additional hurdles posed by them.
In particular, recall that we cannot simply identify regions where all states can obtain the same value (see \cref{ex:changing_secs_in_games}), hence figuring out the optimal infinite play is significantly more involved.
If we were given correct bounds (and regions where the infinite play takes place), we could apply the ideas of \cite{DBLP:conf/cav/AshokCDKM17} to reduce the problem to reachability.

\subsection{Rescaling the Reward}\label{app:MP-rescale}

We restate (and prove) a standard result for mean payoff, mainly for illustrative reasons.
\begin{lemma}\label{lem:addConstantMP}
	Let $\statereward : \allstates \to \Reals$ be a reward function for an SG $\G$.
	For all states $s \in \states$, let $\hat\statereward(s) = \statereward(s) \cdot a + b$ be a rescaled reward function, where $a,b\in \Reals$.
	Then we have $\val_{\G, \meanpay_{\statereward}} \cdot a + b = \val_{\G, \meanpay_{\hat\statereward}}$.
\end{lemma}
\begin{IEEEproof}
	We first show below that for all paths $\rho \in \Infinitepaths<\G>$, it holds that $\meanpay_{\statereward}(\rho) \cdot a + b = \meanpay_{\hat\statereward}(\rho)$. 
	
	Let $\rho \in \Infinitepaths<\G>$ be an arbitrary path.
	\begin{align*}
		\meanpay_{\hat\statereward}(\rho) &= \liminf_{n \to \infty} (\frac{1}{n} \sum_{j=0}^{n-1} \hat\statereward(\rho_j)) \tag{Def.\ of mean payoff}\\
		&= \liminf_{n \to \infty} (\frac{1}{n} \sum_{j=0}^{n-1} \statereward(\rho_j) \cdot a + b) \tag{Def.\ of $\hat\statereward$}\\
		&= 	\left(\liminf_{n \to \infty} (\frac{1}{n} \sum_{j=0}^{n-1} \statereward(\rho_j))\right) \cdot a + b \tag{$a$ and $b$ independent of $j$ and $n$}\\
		&= \meanpay_{\statereward}(\rho) \cdot a + b \tag{By definition of mean payoff}
	\end{align*}
	The proof for the value, which is the expectation over the paths, follows directly from linearity of expectation.
	\begin{align*}
		\val_{\G, \meanpay_{\hat\statereward}} &= {\sup}_\straa~{\inf}_\strab~\expec_{\G, s}^{\straa, \strab}[\meanpay_{\hat\statereward}] \tag{Def.\ of value} \\
		&= {\sup}_\straa~{\inf}_\strab~a \cdot \expec_{\G, s}^{\straa, \strab}[\meanpay_{\statereward}] + b \tag{Linear expectation} \\
		&= \val_{\G, \meanpay_{\statereward}} \cdot a + b \tag{Def.\ of value}
	\end{align*}
	This concludes the proof.
\end{IEEEproof}

This lemma allows us to rescale state rewards, as we can compute the mean payoff with the modified rewards and then obtain the original mean payoff by reversing the rescaling.
This allows us to only consider positive rewards.

\subsection{Bellman Equations for Mean Payoff} \label{app:mp-bellman2}

The mean payoff values are a fixpoint of \cref{eq:bellman}.
Observe that we do not use \cref{eq:bellman} to compute the mean payoff directly, but instead consider the total reward obtained from \cref{eq:bellmanTR}. 

\begin{lemma} \label{lem:MP-fixpoint-eq2}
	Let $\G$ be an SG and $\Phi$ a mean payoff objective.
	Then $\val_{\G, \Phi}$ is a fixpoint of applying \cref{eq:bellman}.
\end{lemma}
\begin{IEEEproof}
	The proof essentially follows from the fact that mean payoff is prefix independent, i.e.\ the value obtained by a run does not change if we add or remove a finite prefix from it.
	Assume for contradiction that $\val_{\G, \Phi}$ does not satisfy the equation, i.e.\ there exists a state $s$ where the optimal expected value of $\val_{\G, \Phi}$ over its successors does not equal its value.
	Moreover, let us assume for simplicity that $s$ is a Maximizer state (the argument for Minimizer is analogous).
	The value of state $s$ can be at most the value over its successors, since in $s$ we can simply pick the optimal action once and then replicate the witness strategies in each of its successors.
	Dually, the value of $s$ can not be larger, since the witness strategy for this value has to achieve, on average, the value in $s$.
	Both of the statements can formally be obtained from the existing theory on mean payoff applied to the Markov chain induced by the witness strategies, see e.g.\ \cite[Thm.~8.2.6]{DBLP:books/wi/Puterman94}, which shows that $\val_{\MC, \Phi}(s) = \sum_{s' \in \allstates} \trans(s, s') \cdot \val_{\MC, \Phi}(s')$ by Eq.~(8.2.11).
	(We remark that \cite{DBLP:books/wi/Puterman94} refers to mean payoff as \emph{gain} and to Markov chains with rewards as \emph{Markov reward process} (MRP).)
\end{IEEEproof}


\subsection{Convergence of Value Iteration} \label{app:mp-convergence}

Here we show that iterating a total reward value iteration according to \cref{eq:bellmanTR} allows us to infer the mean payoff in the limit.
While~\cite[Sec.~5.2]{DBLP:conf/spin/ChatterjeeH08} proposed to extend the reasoning from~\cite[Sec.~2]{DBLP:journals/tcs/ZwickP96} to account for probabilities, we provide a much simpler proof by reducing to MDP and using classical results from \cite{DBLP:books/wi/Puterman94}.

\begin{lemma} \label{stm:mp-convergence}
	Let $\G$ be an SG and $\meanpay_{\statereward}$ a mean payoff objective.
	Let $x_0 : \states \to \RealsNonneg$ assign 0 to every state, and $(x_i)_{i\in\Naturals_0}$ be the sequence of assignments computed by iterating \cref{eq:bellmanTR}, using $\statereward$ as the reward function.
	Then we have
	\begin{equation*}
		{\lim}_{k \to \infty} x_k / k = \val_{\G, \meanpay_{\statereward}}
	\end{equation*}
\end{lemma}
\begin{IEEEproof}
	Recall that there exists a sequence of strategies $\straa_k$ that achieve a $k$-step total reward of at least $x_k$ against any minimizer strategy. These strategies can be obtained by e.g.\ simply following the witnesses of \cref{eq:bellmanTR}.
	For a formal proof, we extend the induction in~\cite[Appendix B.1]{SGreward} to also include $k$-step-bounded optimality of these witnesses, similar to~\cite[Remark 10.104]{DBLP:books/daglib/0020348}.
	Moreover, let $\strab^*$ be a (memoryless, deterministic) mean payoff optimal minimizer strategy in $\G$ and fix $\MDP = \G^{\cdot, \strab^*}$ the induced MDP.

	Assume for contradiction that $\limsup_{k \to \infty} x_k / k > \val_{\G, \meanpay_{\statereward}}$.
	(The dual case for $\liminf$ follows analogously.)
	Clearly, $\straa_k$ achieves a total reward of at least $x_k$ in $\MDP$, too (since $\strab^*$ may be sub-optimal for total reward).
	In other words, the optimal $k$-step total reward in the game $x_k$ is a lower bound on the optimal $k$-step total reward in $\MDP$, denoted by $v_k$.
	By applying \cite[Thm.~9.4.1 b)]{DBLP:books/wi/Puterman94}\footnote{Note that $v_n$ there denotes the optimal $n$-step total reward and $g^*$ the optimal mean payoff in $\MDP$.} to $\MDP$, we get that $\lim_{k \to \infty} v_k / k = \val_{\MDP, \meanpay_{\statereward}}$.
	Together, $\lim_{k \to \infty} v_k / k = \val_{\MDP, \meanpay_{\statereward}} \geq \limsup_{k \to \infty} x_k / k > \val_{\G, \meanpay_{\statereward}}$, i.e.\ there exists a strategy in $\MDP$ that achieves a strictly higher mean payoff against $\strab^*$ than $\val_{\G, \meanpay_{\statereward}}$, contradicting optimality of $\strab^*$.
\end{IEEEproof}

\section{Initializing Value Iteration} \label{app:init-vi}

For lower bounds, we observe that $0$ is correct for all considered objectives:
For reachability and safety this follows immediately, for total reward since we assumed rewards to be non-negative, and for mean payoff we have argued that w.l.o.g.\ we can as well assume that all rewards are non-negative (\cref{lem:addConstantMP}).
To obtain upper bounds, reachability and safety are clearly bounded by $1$, while mean payoff is bounded by the maximal occurring reward $\max_{s \in \allstates} \statereward(s)$.
Total reward is a bit more involved.
We aim to compute a rational number that is greater than the highest possible non-infinite total reward.
To this end, we extend the procedure from~\cite[Sec.~3.3]{ensuring} to SG.
In essence, the idea is as follows:
Let $p_{\min}$ denote the minimum transition probability in the SG and let $s$ be any state with non-infinite value.
Fix optimal strategies $(\straa^\ast, \strab^\ast)$.
Since $s$ has a finite value, the probability to eventually reach a region where zero rewards are obtained is equal to $1$ in $\G^{\straa^\ast, \strab^\ast}$.
From classical observations on Markov chains \cite{DBLP:books/daglib/0020348}, we thus know that there is a probability of at least $p_{\min}^{\abs{\allstates}}$ for this to happen within $\abs{\allstates}$ steps (the smallest possible probability of a path of length $n$).
This path can obtain at most $R_{\max} = \abs{\allstates} \cdot \max_{s \in \allstates} \statereward(s)$ reward.
Hence, the maximum total reward can be bounded by $\sum_{i = 1}^\infty (1 - p_{\min}^{\abs{\allstates}})^i \cdot R_{\max} < \infty$.
In \cite[Sec.~3.3]{ensuring}, more refined ways are presented.

\section{Proofs}

\subsection{Proof of \cref{lem:VIrecommend}} \label{app:recommenderProof}
In this section, we make use of the notion of \emph{fixpoint-linear} objectives, which we introduce in \cref{def:fixpoint_linear}.
Effectively, we require that $\val_{\G, \Phi}(s) = \fpoff_\Phi(s) + \opt_{a \in \act(s)} \val_{\G, \Phi}(s, a)$, i.e.\ the value function satisfies an (affine) linear equation.

We repeat the notion of \emph{safe} strategies, which are those that take actions optimal w.r.t.\ the value.
\begin{definition}[\cite{DBLP:conf/isaac/AnderssonM09}]
	A strategy profile $\pi = (\straa, \strab)$ is \emph{safe} (for $\Phi$) if $\pi(s) \in \argopt_{a \in \act(s)} \val_{\G, \Phi}(s, a)$ for all states $s \in \allstates$.
\end{definition}
Following safe strategies for any finite time is correct:
\begin{lemma} \label{stm:safe_strategy_finite_horizon}
	Suppose $\Phi$ is fixpoint-linear, $\pi$ a safe strategy for $\Phi$, and $\pi^*$ an optimal strategy.
	Following $\pi$ for $k$ steps and then $\pi^*$ yields optimal values.
\end{lemma}
\begin{IEEEproof}
	We proceed by induction over $k$, showing the statement for all states of $\G$.
	Clearly, the statement holds for $k = 0$ -- we simply follow an optimal strategy.
	From $k$ to $k + 1$, fix a state $s$.
	By assumption, $\val_{\G, \Phi}(s) = \fpoff_\Phi(s) + \opt_{a \in \act(s)} \val_{\G, \Phi}(s, a)$.
	Since $\pi$ is safe, $\pi(s)$ is such a witness, i.e.\ $\val_{\G, \Phi}(s) = \fpoff_\Phi(s) + \val_{\G, \Phi}(s, \pi(s))$.
	By definition, $\val_{\G, \Phi}(s, \pi(s)) = \sum_{s' \in \allstates} \trans(s, \pi(s), s') \cdot \val_{\G, \Phi}(s')$.
	Inserting the induction hypothesis for $\val_{\G, \Phi}(s')$ yields that following $\pi$ for $k$ steps and then $\pi^*$ is optimal.
	Thus, following $\pi$ for $k+1$ steps in $s$ is optimal, too, concluding the proof.
\end{IEEEproof}
Unfortunately, following a safe strategy \emph{forever} may not be correct.
In other words, \enquote{safety} is only necessary but not sufficient for optimality.
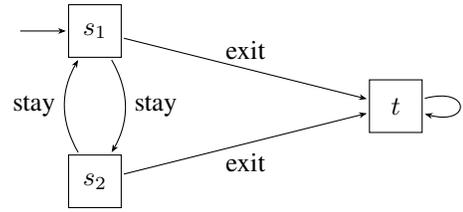
\begin{figure}[t]
	\centering
	\begin{tikzpicture}
		\drawdummy (init) at (0,0) {};
		\drawbox (p) at (1,0) {$s_1$};
		\drawbox (q) at (1,-2) {$s_2$};
		\drawbox (1) at (5,-1) {$t$ };
		
		\draw[->] (init) to (p);
		\draw[->]  (p) to[bend left] node [midway,anchor=west] {$\text{stay}$}(q) ;
		\draw[->]  (q) to [bend left] node [midway,anchor=east] {$\text{stay}$} (p);
		\draw[->]  (p) to node [midway,anchor=south] {$\text{exit}$} (1);
		\draw[->]  (q) to node [midway,anchor=north] {$\text{exit}$} (1);
		\draw[->]  (1) to [loop right] node [midway,anchor=west] {} (1);
	\end{tikzpicture}
	\caption{
		An example of an SG (also MDP).
		Considering the reachability query with target $\targets = \{t\}$, following \enquote{safe} actions is not necessarily an optimal strategy.
	}
	\label{fig:strat-CE}
\end{figure}
\begin{example}\label{ex:strat-CE}
	Consider the system in \cref{fig:strat-CE} (which is an MDP).
	We consider reachability with target $\targets = \{t\}$.
	Clearly, all states have a value of $1$.
	However, playing the action \enquote{stay} in both $s_1$ and $s_2$ reduces their reachability probability to $0$.
	Thus, simply picking any optimal action $a \in \argopt_{a \in \act(s)} \val_{\G, \Phi}(s, a)$ is not guaranteed to yield a correct strategy.
	Observe that even the value iterates have $x_i(s) = 1$ for all states $s$ and $i \geq 2$.
	Consequently, we cannot choose an optimal action solely based on the current iterates, either.
\end{example}
We need to ensure that strategies additionally make \enquote{progress} instead of being stuck in spurious fixpoints.
This is, for example, ensured by our proposed construction of only switching action if there is a strictly better alternative, as we show in the following.
Alternatively, one can choose among optimal actions uniformly at random, which also ensures that a \enquote{progressing} action is chosen with non-zero probability.
For a slightly different perspective, we direct the interested reader to \cite{DBLP:conf/isaac/AnderssonM09}, which ensures progress by choosing actions minimizing the distance to the goal, using an \enquote{attractor} computation.
(In the context of reachability, \cite{DBLP:conf/isaac/AnderssonM09} identifies \enquote{stopping} as additional criterion to be satisfied on top of safety, capturing this idea of progressing towards the target.)

Before proceeding with the main proof, we show a useful auxiliary statement for a broad variety of objectives.
Intuitively, we show that strategies produced by \cref{lem:VIrecommend} make progress towards \enquote{the right} BSCCs.
\begin{lemma} \label{stm:no_stupid_ecs}
	Let $\mathsf{VI}$ a Bellman update of the following form:
	We have $\mathsf{VI}(x)(s) = c(s)$ for \emph{special states} $s \in C$ which are all absorbing, with $c(s) \geq 0$ a constant.
	For other states, we have $\mathsf{VI}_\Phi(x)(s) = \statereward(s) + \opt_{a \in \act(s)} x(s, a)$ with $\statereward(s) \geq 0$.
	Moreover, suppose that the iterates $x_i$, i.e.\ $x_{i+1} = \mathsf{VI}_\Phi(x_i)$, are either (i)~monotonically increasing or (ii)~monotonically decreasing.

	Let $\pi_i = (\straa_i, \strab_i)$ be the strategies obtained in step $i$ as defined by \cref{lem:VIrecommend}.
	Then, every BSCC $R$ in $\G^{\pi_i}$ either exists in all $\G^{\pi_j}$ with $j \leq i$ or $\statereward(s) > 0$ for some $s \in R$.
\end{lemma}

\begin{IEEEproof}
	Suppose a new BSCC $R$ emerges in step $k$.
	Since special states are absorbing and never change their value, all states in this BSCC are not in $C$.
	Thus, there exists a state $s \in R$ where one player changed strategy due to a \emph{strict} improvement.
	In other words, $x_{k-1}(s) \prec_s \statereward(s) + x_{k-1}(s, \pi_k(s)) = x_k(s)$.
	(The important fact is that $x_{k-1}(s) \neq x_k(s)$.)
	Moreover, $(s, \pi_{k-1}(s))$ exits the EC corresponding to $R$, i.e.\ $(R, \{\pi_k(s') \mid s' \in R\})$ (otherwise $R$ would already exist in $\G^{\pi_{k-1}})$.

	We first treat the case of monotonically increasing $x_i$, i.e.\ $x_i \leq x_{i+1}$ (which is the case for reachability, total reward, and mean payoff).
	Then, we have
	\begin{align*}
		x_{k}(s) & = \mathsf{VI}(x_{k-1})(s) \\
			& = \statereward(s) + {\sum}_{s' \in R} \trans(s, \pi_k(s), s') \cdot x_{k-1}(s') \\[2ex]
			& = \begin{aligned}
				\statereward(s) & + \trans(s, \pi_k(s), s) \cdot x_{k-1}(s) +                            \\
				                & {\sum}_{s' \in R, s' \neq s} \trans(s, \pi_k(s), s') \cdot x_{k - 1}(s')
			\end{aligned} \\[2ex]
			& \leq \begin{aligned}
				\statereward(s) & + \trans(s, \pi_k(s), s) \cdot x_{k-1}(s)                           \\
				                & + {\sum}_{s' \in R, s' \neq s} \trans(s, \pi_k(s), s') \cdot x_k(s').
			\end{aligned}
	\end{align*}
	We can repeatedly insert the definition of $x_k(s') = \statereward(s') + x_{k-1}(s', \pi_k(s'))$, split by the probability of transitioning from $s'$ to $s$ and to other states $s''$, and again use the monotonicity of $x_i$, i.e.\ $x_{k-1}(s'') \leq x_{k}(s'')$.
	In other words, we consider all potential finite paths from $s$ back to itself inside this EC, which in sum have a probability of 1.
	Formally, let $P = \{\rho \mid s_1 = s \land s_n = s \land s \neq s_i \text{ for all $1 < i < n$}\}$ and set $\statereward(\rho) = \sum \statereward(\rho_i)$ the sum of rewards along a path.
	Then, by above reasoning,
	\begin{align*}
		x_{k}(s) & \leq {\sum}_{\rho \in P} {\Pr}_{\G, s}^{\pi_i}[\rho] \cdot (\statereward(\rho) + x_{k-1}(s))             \\
		         & = \left({\sum}_{\rho \in P} {\Pr}_{\G, s}^{\pi_i}[\rho] \cdot \statereward(\rho)\right) + x_{k-1}(s).
	\end{align*}
	We end up with $x_k(s) \leq W + x_{k-1}(s)$, where $W$ is the weighted sum of all rewards obtained along these paths.

	Observe that $W = 0$ if and only if $\statereward(s) = 0$ for all $s \in R$.
	Thus, when $W = 0$, then $x_k(s) \leq x_{k-1}(s)$ and, by monotonicity, $x_{k-1}(s) \leq x_k(s)$, a contradiction to $x_{k-1}(s) \neq x_k(s)$.
	In summary, we can only have a BSCC if $\statereward(s) > 0$ or this BSCC already exists from step $0$ (i.e.\ we never switched action).

	For the case of monotonically decreasing $x_i$ (i.e.\ safety), the above reasoning still applies.
	By considering the paths inside $R$, we dually get $x_k(s) \geq W + x_{k-1}(s)$.
	By monotonicity, $x_k(s) \leq x_{k-1}(s)$.
	Thus, if $W = 0$, $x_k(s) = x_{k-1}(s)$, contradicting strict improvement.
\end{IEEEproof}
We remark that a similar statement can be derived when choosing uniformly at random among all optimal actions.
Then, we only get BSCCs if \emph{all} exiting actions are strictly worse.
However, intuitively, this cannot happen if $\statereward(s) = 0$ inside the BSCC, since a strictly better value cannot appear \enquote{out of thin air}.

With these tools at hand, we are ready to prove \cref{lem:VIrecommend}, which we restate here for readability.
\restatevirecommender*
\begin{IEEEproof}
	For readability, we write $\val$ instead of $\val_{\G, \Phi}$.

	\underline{Strategies are safe}:
	First, we show that eventually the actions picked by either strategy are safe, i.e.\ there exists $k_0$ such that $\pi_k(s) \in \argopt_{a \in \act(s)} \val(s, a)$ for all $k \geq k_0$.
	(Note that this does \emph{not} imply that the strategies remain constant.)
	This follows immediately from the convergence guarantees for reachability and safety~\cite[Section 3]{DBLP:conf/spin/ChatterjeeH08}, as well as total reward~\cite[Appendix B.1]{SGreward}.
	There, we know that $\lim_{i \to \infty} x_i(s) = \val(s)$.
	Thus, suppose that $\pi_i(s)$ would pick a suboptimal action $b$ infinitely often.
	This means that $x_i(s, a^*) \prec x_i(s, b)$ for all optimal actions $a^* \in \argopt_{a \in \act(s)} \val(s, a)$.
	However, since $x_i(s, b) \to \val_{\G, \Phi}(s, b)$ by the convergence guarantees, we obtain a contradiction.

	In the case of mean payoff, we need to argue separately, since the iterates do not directly yield the value.
	We employ reasoning similar to the one in the proof of \cref{stm:mp-convergence}.
	Suppose that a suboptimal action $b$ is selected infinitely often.
	This means that by using action $b$, we can obtain at least as much $k$-step total reward as by using any optimal action $a^*$ for arbitrarily large $k$.
	(This follows from results on total reward iteration.)
	Thus, $\limsup_{k \to \infty} \val_{\G, \mathsf{TR}}(s, b) / k \geq \val(s, a^*) = \val(s)$.
	Now, observe the left hand side converges to the optimal mean payoff achieved under $b$ (see \cref{stm:mp-convergence} for further details), contradicting the sub-optimality of $b$.

	\underline{Strategies do not follow spurious fixpoints}:
	Now, we know that strategies are safe.
	However, this does not exclude the issues outlined in \cref{ex:strat-CE}.
	Using \cref{stm:no_stupid_ecs}, we now show that in all BSSCs induced by these strategies they eventually do actually obtain the optimal value.
	Formally, let $R$ a BSCC in $\G^{\pi_i}$. 
	We show that $\val_{\G^{\pi_i}, \Phi}(s) = \val_{\G, \Phi}(s)$ for all $s \in R$ and large enough $i$.
	We consider each objective separately.

	We begin with reachability.
	We prove that $\pi_i$ eventually only induces BSCCs where either all states have value zero or equal a target state.
	The claim then follows immediately.
	By \cref{stm:no_stupid_ecs}, the only non-target BSCCs that can exist are those which exist from the beginning.
	Thus, let $R$ a BSCC in $\G^{\pi_0}$ and $R' = \{s \mid s \in R \land \val(s) > 0\} \subseteq R$ the states with non-zero value.
	We argue that a state $s$ in $R'$ eventually switches action.
	We know that the iterates $x_i$ converge to the true value (which is $> 0$ in all states of $R'$).
	Consequently, there has to be a step $k$ where for the first time any state of $R'$ changes from $x_{k-1}(s) = 0$ to $x_k(s) > 0$ (this could happen for multiple states at once).
	However, this necessarily implies that state $s$ changes action:
	By assumption, at step $k - 1$, all states $s' \in R$ have a value of $x_{k-1}(s') = 0$, consequently $\sum_{s' \in R} \trans(s, \pi_{k-1}(s), s') \cdot x_{k-1}(s') = 0$.
	Thus, the witness action necessarily needs to exit $R$ with positive probability.
	In summary, for large enough $i$, there are no non-target BSCCs under $\pi_i$ where $\val(s) > 0$.
	For safety, the same reasoning can be applied.
	(For these two objectives, is is known that this already happens within at most $\abs{\allstates}$ steps.)

	For total reward, we show that eventually all BSCCs $R$ in $\G^{\pi_i}$ have $\val(s) = 0$ for all states $s \in R$.
	To this end, suppose that the $\pi_i$ we consider already are safe.
	Let $R$ be a BSCC of $\G^{\pi_i}$.
	Suppose that $\statereward(s) > 0$ for some state $s \in R$.
	By \cref{stm:safe_strategy_finite_horizon}, both players are content with remaining in this BSCC arbitrarily long.
	This includes visiting $s$ arbitrarily often, implying that the value in this BSCC is infinite (otherwise, Minimizer would eventually leave).
	However, this is excluded by assumption.
	Now, we follow the same reasoning as above:
	If there initially is a BSCC with states of non-zero value, this BSCC eventually dissolves.
	Together, the only BSCCs that remain are those with both state reward zero and a value of zero, proving the claim.

	Finally, for mean payoff, this immediately follows from the reasoning of \cref{stm:mp-convergence}:
	All actions which are $k$-step total reward optimal for large enough $k$ are mean payoff optimal.

	\underline{Strategies behave correctly on transient states}:
	We have shown that for large enough $i$ the strategies $\pi_i$ \enquote{do the correct thing} once they reach BSCCs.
	It remains to argue that we do not miss out on anything on the way.
	To prove this, first observe that eventually the strategies $\pi_i$ we consider are safe by above reasoning.
	Moreover, on all induced BSCCs the strategy $\pi_i$ is optimal.
	Thus, there exists an optimal strategy $\pi^*$ which behaves like $\pi_i$ on all BSCCs induced by $\pi_i$.
	Formally, let $\mathcal{B}_i$ the set of all BSCCs in $\G^{\pi_i}$.
	Then, we have $\pi_i(s) = \pi^*(s)$ for all $s \in R, R \in \mathcal{B}_i$.
	Next, observe that the probability to reach BSCCs in finitely many steps approaches 1, i.e.\ $\lim_{k \to \infty} \Pr_{\G^{\pi_i},s}[\reach^{\leq k} \mathcal{B}_i] = 1$.
	Thus, we apply \cref{stm:safe_strategy_finite_horizon} for arbitrarily large $k$ to $\pi_i$ and $\pi^*$.
	By the above reasoning, the probability not to be in a BSCC after $k$ steps (i.e.\ switching from $\pi_i$ to $\pi^*$) tends to zero, thus the value achieved by the strategy composed of $\pi_i$ and $\pi^*$ after $k$ steps tends to the value achieved by $\pi_i$.
	Formally, set $\pi_i^k$ the strategy following $\pi_i$ for $k$ steps and then $\pi^*$ as defined above.
	Then,
	\begin{equation*}
		\val_{\G^{\pi_i^k},\Phi}(s) = R_i^k + {\sum}_{s' \in \allstates} {\Pr}_{\G, s}^{\pi_i^k}[\rho_k = s'] \cdot \val_{\G^{\pi^*}, \Phi}(s'),
	\end{equation*}
	where $R_i^k$ are the weighted rewards obtained by $\pi_i$ in the first $k$ steps (which is optimal by \cref{stm:safe_strategy_finite_horizon}) and $\Pr_{\G, s}^{\pi_i^k}[\rho_k = s']$ denotes the probability of ending up in $s'$ after $k$ steps.
	(This can also be directly obtained by simply considering all finite paths of length $k$.)
	To conclude, observe that by previous argument for large enough $k$ the probability of ending up in a BSCC approaches 1, where $\val_{\G^{\pi_i}, \Phi} = \val_{\G^{\pi^*}, \Phi} = \val_{\G, \Phi}$.
\end{IEEEproof}
\subsection{Proof of \cref{stm:secs_are_without_reward}} \label{app:proof_secs_without_reward}

\restatesecswithoutreward*

\begin{IEEEproof}
	Fix $E$ and $f$ as in the assumptions.
	Choose an arbitrary pair of states $s$ and $t$.
	We show that $f(s) = f(t)$.
	Since $E$ is an EC, there exists a pair of memoryless deterministic strategies that reach $t$ almost surely, using only actions of $E$.
	Fix such a strategy pair $\pi = (\straa, \strab)$ (these strategies do not need to be optimal for $\Phi$).
	Then $f(s) = \fpoff_\Phi(s) + \trans(s, \pi(s), t) \cdot f(t) + \sum_{s' \in R \setminus \{t\}} \trans(s, \pi(s), s') \cdot s'$ with $\fpoff_\Phi(s) \geq 0$.
	Since the state $t$ is reached almost surely, by repeatedly inserting this equation for $f(s')$ we end up with $f(s) = V + f(t)$, where $V \geq 0$.
	More formally, for a path $\rho = s_1 a_1 \cdots a_{n-1} s_n$, set $\fpoff_\Phi(\rho) = \sum_{i = 1}^n \fpoff_\Phi(s_i)$ the sum of all rewards obtained along $\rho$ and $\trans(\rho) = \prod_{i = 1}^{n - 1} \trans(s_i, a_i, s_{i+1})$ its probability.
	Define $P = \{\rho \mid s_1 = s \land s_n = t \land t \neq s_i \text{ for all $i < n$}\}$ the set of all paths from $s$ to $t$.
	Then, $f(s) = \sum_{\rho \in P} \trans(\rho) \cdot (\fpoff_\Phi(\rho) + f(t))$.
	We have that $\sum_{\rho \in P} \trans(\rho) = 1$ and $\fpoff_\Phi(\rho) \geq 0$ for all $\rho$.
	Together, we get the desired result, namely that $f(s) = f(t) + \sum_{\rho \in P} \trans(\rho) \cdot \fpoff_\Phi(\rho) \geq f(t)$.
	We can apply the same idea to \enquote{get back} to $s$ and obtain $f(s) = V + V' + f(s)$, showing that $V = V' = 0$ and thus $\fpoff_\Phi(s) = 0$ (note that we exclude $f(s) = \infty$).
	As an immediate consequence, we also get that $f(s) = f(t)$.
\end{IEEEproof}

\subsection{Proof of \cref{stm:secs_assignment_equal}} \label{app:proof_assignment_equal}

\restatesecsequality*

\begin{IEEEproof}
	Note that the backward direction follows immediately.
	The forward direction follows exactly as in \cref{stm:secs_are_without_reward}.
	Since $E$ is SEC for $f$, we can again follow a \enquote{loop} from state $s$ to $t$ to derive that $\fpoff_\Phi(s) = 0$, which implies $f(s) = f(t)$.
\end{IEEEproof}
\subsection{Proof of \cref{lem:car_upgraded}} \label{app:proof_car}

\restatecar*

We prove a stronger statement, implying the above.

\begin{lemma} \label{lem:car_real}
	Let $f$ a fixpoint for $\fixpointup_\Phi$.
	Set $d(s) = f(s) - \val_{\G, \Phi}(s)$ its difference to the value and $d^+ = \max_{s \in S} d(s)$ and $d^- = \min_{s \in S} d(s)$ the maximum and minimum difference.
	There exist $\Phi$-SECs $E^+$ and $E^-$ with $d(s) = d^+$ and $d(s) = d^-$ for all states in $E^+$ and $E^-$ respectively.
\end{lemma}

\begin{IEEEproof}
	For readability, we write $\val$ instead of $\val_{\G, \Phi}$.
	Let $f$ as in the assumptions.
	Observe that $d(s, a) = \sum_{s' \in S} \trans(s, a, s') d(s) = \sum_{s' \in S} \trans(s, a, s') (f(s') - \val(s')) = f(s, a) - \val(s, a)$.
	Let $d^+ = \max_{s \in S} d(s)$ the maximal difference between $f$ and $\val$.
	Define $S^+ = \{s \mid d(s) = d^+\}$ all states witnessing the maximal difference.
	(Note that $d^+$ might be zero, e.g.\ if $f \equiv \val$, or even smaller than zero if $f < \val$.)

	We show that $S^+$ necessarily contains at least one SEC for $f$.
	In particular, we prove that for every $s \in S^+$, there exists an action $a \in \actions(s)$ such that (i)~$(s, a)$ does not exit $S^+$ and (ii)~$a$ is a witness for the SEC condition.
	Thus, fix an arbitrary state $s \in S^+$.
	Observe that $d(s, a) = \sum_{s' \in S} \trans(s, a, s') (f(s') - \val(s'))$.
	By definition of $d^+$, we have $f(s) - \val(s) \leq d^+$ for all states $s$.
	Thus, the above sum is strictly smaller than $d^+$ exactly if at least one successor $s'$ of $(s, a)$ has $d(s') < d^+$, meaning $s' \notin S^+$.
	In other words, $(s, a)$ exits $S^+$ iff $d(s, a) < d^+$.
	We proceed by a case distinction on the owner of $s$.
	
	\underline{Maximizer}:
	By definition, $\val$ is a fixpoint for $\fixpointup_\Phi$, i.e.\ $\val(s) = \fpoff_\Phi(s) + \max_{a \in \act(s)} \val(s, a)$.
	For every action $a' \in \actions(s)$, we thus have $\val(s) \geq \fpoff_\Phi(s) + \val(s, a')$:
	The action $a'$ can at most be the optimal one in the maximum.
	Together, for every action $a' \in \actions(s)$ we have $d(s) = f(s) - \val(s) \leq f(s) - \val(s, a') - \fpoff_\Phi(s)$.
	Recall that $f$ is $\fixpointup_\Phi$ fixpoint; let $a_s$ the witness action for $f(s) = \fpoff_\Phi(s) + \max_{a \in \act(s)} f(s, a)$, i.e.\ $f(s) = \fpoff_\Phi(s) + f(s, a_s) = \fixpointup_\Phi(f, s)$.
	Suppose that $(s, a_s)$ exits $S^+$.
	By the above argument, we have $d(s, a_s) < d^+$.
	Since $s \in S^+$, we have $d(s) = d^+$.
	Together, $d(s, a_s) < d(s) \leq f(s) - \val(s, a_s) - \fpoff_\Phi(s) = f(s, a_s) - \val(s, a_s) = d(s, a_s)$, a contradiction.
	Thus, $(s, a_s)$ does not exit $S^+$.
	Moreover, by choice of $a_s$ we have $f(s) = f(s, a_s) = \fixpointup_\Phi(f, s)$.

	\underline{Minimizer}:
	Since $f(s) = \fpoff_\Phi(s) + \min_{a \in \actions(s)} f(s, a)$, we analogously get $d(s) = f(s) - \val(s) \leq \fpoff_\Phi(s) + f(s, a') - \val(s)$ for any $a' \in \act(s)$.
	As above, let $a_s$ such that $\val(s) = \fpoff_\Phi(s) + \val(s, a_s)$ and assume $(s, a_s)$ exits $S^+$.
	Then $d(s, a_s) < d(s) \leq f(s, a_s) - \val(s, a_s) = d(s, a_s)$, again a contradiction.

	Quite surprisingly, even though $a_s$ was chosen as arbitrary witness for $\val(s)$, it turns out that $f(s, a_s) = \fixpointup_\Phi(f, s)$, too:
	First, observe that $f(s) = \val(s) + d^+$ for all states in $S^+$.
	Since $(s, a_s)$ does not exit $S^+$, this in particular holds for all successors of $(s, a_s)$.
	Moreover, clearly $f(s) = \fpoff_\Phi(s) + \min_{a \in \act(s)} f(s, a) \leq \fpoff_\Phi(s) + f(s, a_s)$.
	Together, we get
	\begin{align*}
		f(s) & = \val(s) + d^+ = \fpoff_\Phi(s) + \val(s, a_s) + d^+ \\
			& = \fpoff_\Phi(s) + \left({\sum}_{s' \in S} \trans(s, a_s, s') \val(s')\right) + d^+ \\
			& = \fpoff_\Phi(s) + {\sum}_{s' \in S} \trans(s, a_s, s') (\val(s') + d^+) \\
			& = \fpoff_\Phi(s) + {\sum}_{s' \in S} \trans(s, a_s, s') f(s') \\
			& = \fpoff_\Phi(s) + f(s, a_s).
	\end{align*}
	In short, $\fixpointup_\Phi(f, s) = f(s) = f(s, a_s) + \fpoff_\Phi(s)$.

	Together, every state has an action $a_s$ that (i)~does not exit $S^+$ and (ii)~witnesses $f(s) = \fpoff_\Phi(s) + \opt^s_{a \in \act(s)} f(s, a) = \fixpointup_\Phi(f, s)$.
	To conclude, consider the game where we restrict to $S^+$ and set $\act(s) = \{a_s\}$, which is well defined by (i).
	This game contains at least one EC, with $\fpoff_\Phi(s) = 0$ by \cref{stm:secs_are_without_reward}, and, by (ii), this is an SEC in $\G$.
	Moreover, since this SEC is a subset of $S^+$, we have that all states in SEC have $d(s) = d^+$, i.e.\ this EC qualifies for $E^+$.

	The proof for $d^-$ and $E^-$ follows analogously by \enquote{exchanging} the proofs for Maximizer and Minimizer.
\end{IEEEproof}
\subsection{Proof of \cref{lem:inflate_deflate}} \label{app:proof_inflate_deflate}

\restateinflatedeflate*

\begin{IEEEproof}
	We prove the Maximizer case, the Minimizer follows analogously.
	Since $E$ is SEC, we know that all states in $E$ obtain the same value in $\G$ by \cref{stm:secs_are_without_reward}.
	Let $\strab^*$ an optimal Minimizer strategy.
	Then, all states of $E$ also have the same optimal value in the induced MDP $\G^{\cdot, \strab^*}$, too.


	Now, suppose an optimal Maximizer strategy $\straa^*$ remains in the EC, i.e.\ the probability to reach an state outside of $E$ from $s$ under $(\straa^*, \strab^*)$ is 0.
	Then, we have $\val_{\G, \Phi}(s) = \stayVal_{\G^{\cdot, \strab^*}}(E, s)$.
	Using the intuition of \cref{eq:mdp_view_intuition}, we get $\stayVal_{\G^{\cdot, \strab^*}}(E, s) \leq \stayVal_{\G^{\cdot, \strab}}(E, s)$ for an arbitrary Minimizer strategy $\strab$.
	For the second part, suppose $\straa^*$ leaves the EC with non-zero probability at some state $\hat{s}$.
	The value obtained in $\hat{s}$ by leaving is at most as large as the best exit $\bE^\Box_{\val_{\G, \Phi}}(E)$ by definition of best exit, i.e.\ $\val_{\G, \Phi}(\hat{s}) \leq \bE^\Box_{\val_{\G, \Phi}}(E)$.
	Since $f$ is an upper bound on the value, we also get $\bE^\Box_{\val_{\G, \Phi}}(E) \leq \bE^\Box_f(E)$.
	Consequently, $\val_{\G, \Phi}(\hat{s}) \leq \bE^\Box_f(E)$.
	Now, to transfer the result from the exiting state $\hat{s}$ to all other states, we crucially require the SEC assumption and, in particular, \cref{stm:secs_assignment_equal}.
	This yields $\val_{\G, \Phi}(s) = \val_{\G, \Phi}(\hat{s})$.

	Together, we get that $\val_{\G, \Phi}(s) \leq \stayVal_{\G^{\cdot, \strab}}(E, s)$ in case an optimal Maximizer strategy remains inside the EC and $\val_{\G, \Phi}(s) = \val_{\G, \Phi}(\hat{s}) \leq \bE^\Box_f(E)$ if an optimal Maximizer strategy leaves.
	This proves the claim.
\end{IEEEproof}
\subsection{Proof of \cref{stm:algo_sg_correct}} \label{app:proof_sg_correct}

\restatesgcorrect*

\begin{IEEEproof}
	\underline{Correctness:}
	We prove that $\lb_i \leq \val \leq \ub_i$ by induction.
	The bounds are updated in three places: (i)~the initialization (\cref{line2:initbounds}), (ii)~the updates (\cref{line2:BU-L,line2:BU-U}), and (iii)~the inflating and deflating (\cref{line2:defl-upd,line2:infl-upd}).

	The bounds are correct after initialization by assumption, see \appref{app:init-vi}.
	For the updates of the upper bounds (lower bounds follow analogously), observe that even updating with the optimal choice, i.e.\ setting
	\begin{multline*}
		\ub_i(s) = \fixpointup_\Phi(\ub_{i-1}, s) = \\
		v(s) + \opt_{a \in \actions(s)} {\sum}_{s' \in \allstates} \trans(s, a, s') \cdot \ub_{i-1}(s') \geq \\
		\fixpointup_\Phi(\val_{\G, \Phi}),
	\end{multline*}
	is correct by linearity of the objective and the induction hypothesis $\ub_{i-1} \geq \val_{\G, \Phi}$.
	See also \cite{DBLP:conf/spin/ChatterjeeH08,SGreward} for objective-specific reasoning. 
	Since we fix an arbitrary choice of the Minimizer for the update, $\ub_i(s) \geq \fixpointup_\Phi(\ub_{i-1}, s) \geq \val_{\G, \Phi}(s)$ in all Minimizer states.
	For the correctness of deflating and inflating, observe that we can directly apply \cref{lem:inflate_deflate}.
	Together, we obtain correctness, i.e.\ $\lb_i \leq \val \leq \ub_i$ for all $i$.
	\medskip
	
	\underline{Termination:}
	Similar to the proof of \cref{alg:MDP-view}, this proof relies on the strategy recommender to eventually give optimal strategies $(\straa^\ast, \strab^\ast)$.
	Moreover, for the local algorithm we require the strategy recommender to be stable.
	Note that the recommended strategies may change infinitely often, we only know that eventually each recommended strategy is optimal.
	Thus, the fixpoint updates as well as deflating and inflating steps eventually are only applied to \enquote{optimal MDPs} $\G^{\cdot, \strab^\ast}$ respectively $\G^{\straa^\ast, \cdot}$.

	It remains to show that the repeated updates of upper bounds together with deflating converge to the correct solution.
	The dual statement for inflating follows analogously.
	Since there are only finitely many memoryless deterministic strategies, there exists an optimal Minimizer strategy $\strab^\ast$ which is recommended infinitely often.
	Let $k_i$ denote the sequence of occurrences of $\strab^\ast$.

	Since $\ub_i(s)$ is monotonically decreasing and bounded from below by $\val_{\G, \Phi}(s)$ (as shown in the correctness proof), it converges to some value $\ub(s) = \lim_{i \to \infty} \ub_i(s) \geq \val_{\G, \Phi}(s)$.
	(Note that this limit is potentially never attained during the execution of the algorithm.)
	By definition of $\ub_i(s)$, we necessarily have that $\ub(s)$ is a fixpoint of $\fixpointup_\Phi^{\G^{\cdot, \strab^\ast}}$, i.e.\ $\fixpointup_\Phi$ applied in the MDP.
	Let $\straa^*$ be the Maximizer strategy which takes all actions that are witnesses for this fixpoint uniformly at random, i.e.\ for each Maximizer state $s \in \maxstates$ the strategy takes all actions $a \in \actions(s)$ which achieve the optimal value $\fixpointup_\Phi(\ub, s)$.
	Define the induced MC $\MC^* = \G^{\straa^*, \strab^*}$.
	By construction, $\ub$ also is a fixpoint of $\fixpointup_\Phi^{\MC^*}$.
	Fix an arbitrary BSCC $B$ of $\MC^*$.
	This BSCC corresponds to an EC $E$ in $\G^{\cdot, \strab^\ast}$, on which $\ub$ is a fixpoint of $\fixpointup_\Phi$ in $\G^{\cdot, \strab^\ast}$.
	Thus, $E$ is SEC in $\G^{\cdot, \strab^\ast}$ for $\ub$ and $\ub(s) = \ub(t)$ for all states $s, t \in E$.
	Moreover, in every state each exiting action has value \emph{strictly} smaller than all staying action w.r.t.\ $\ub$ (otherwise $\straa^\ast$ would choose this action, too), i.e.\ $\bE^\Box_\ub(E) < \ub(s)$ for all $s$.
	This also means that $E$ is inclusion maximal:
	There is no action that could be added without violating the SEC condition.
	As $\ub$ is the limit of $\ub_i$ and the best exit of $E$ under $\ub$ is strictly smaller than the staying actions of $E$, there necessarily exists a step $i$ where the same holds for the best exit under $\ub_i$.
	Moreover, since $\strab^*$ is optimal, we have that $\stayVal_{\G^{\cdot, \strab^*}, \Phi}(E, s) \leq \val_{\G, \Phi}(s)$.
	Since $E$ is MSEC in the MDP $\G^{\cdot, \strab^\ast}$, it is repeatedly deflated by the algorithm.
	Together, if $\val_{\G, \Phi}(s) < \ub(s)$ for some state $s$ in $E$, we in particular have that $\ub_i(s) > \max(\stayVal_{\G^{\cdot, \strab^*}, \Phi}(E, s), \bE^\Box_{\ub_i}(E))$, contradicting deflation.
	In summary, we eventually have that $\ub_i(s) = \val_{\G, \Phi}(s)$ for all BSCCs of $\MC^*$.

	It remains to argue the same about the other states.
	For every $i$, set $\straa_i$ the Maximizer strategy which takes all $\ub_{k_i}$-optimal actions uniformly at random. 
	There are only finitely many different strategies $\straa_i$ (a strategy choosing actions uniformly at random is uniquely characterized by the set of actions it chooses in each state).
	Hence let $\hat{\straa}$ now equal such a strategy that appears infinitely often in that sequence.
	The upper bounds $\ub_i$ are updated infinitely often according to the transition structure of $\G^{\hat{\straa}, \strab^*}$ by choice of $\straa_i$ and $\hat{\straa}$, i.e.\ $\ub_{k'_i+1}(s) \leq \fixpointup_\Phi^{\G^{\hat{\straa}, \strab^*}}(\ub_{k'_i}, s)$ where $k'_i$ are the indices of occurrences of $\hat{\straa}$.
	By applying a standard absorption argument, $\ub$ necessarily is a fixpoint of $\fixpointup_\Phi^{\G^{\straa_i, \strab^*}}$.
	Moreover, the BSCCs of $\G^{\hat{\straa}, \strab^*}$ necessarily all are subsets of BSCCs of $\MC^*$, by the previous argument the values on these BSCCs equal the true values, and by classical results on Markov chains we get convergence to the true value on all remaining states:
	By applying the same arguments as before (considering all paths of finite length $k$ and letting $k$ tend to infinity), we observe that for fixpoint-linear objectives there is a unique fixpoint on Markov chains once values on all BSCCs are fixed.
	Intuitively, since we eventually reach BSCCs with probability one, the unique solution of the fixpoint equations is given by the values on BSCCs, weighted by the probability of reaching them, together with the weighted sum of all transient paths (which is uniquely determined).
	Since we have a unique fixpoint, and value iteration is guaranteed to converge to that unique fixpoint for all objectives, we obtain the result.
\end{IEEEproof}

\section{Strategy Iteration Variant of \cref{alg:MDP-view}} \label{app:SIversion}

\begin{algorithm}[!h]
	\caption{Generic strategy iteration anytime algorithm} \label{alg:SI}
	\begin{algorithmic}[1]
		\Require SG $\G$, Objective $\Phi$, initial state $\initstate$, and precision $\varepsilon$
		\Ensure $(\lb, \ub)$ such that $\lb \leq \val \leq \ub$ and $\ub(\initstate) - \lb(\initstate) \leq \varepsilon$
		\Commentline{Initialization}
		\State $\lb_0(\cdot) \gets 0$, $\ub_0(\cdot) \gets \infty$ \Comment{Lower and upper bounds}
		\State $\straa \gets \text{arbitrary Maximizer strategy}$
		\State $i \gets 0$
		\Repeat
		\State $i \gets i + 1$
		\medskip
		\Statex \Commentline{Recommender procedure}
		\State $x \gets \val(\G^{\straa,\cdot})$\label{line:si-mdp}
		\State $\straa, \strab \gets$ best-effort strategies inferred from $x$\label{line:si-infer}
		\medskip
		\Statex \Commentline{Compute bounds}
		\State $\lb_{i} \gets x $\label{line:si-l}
		\State $\ub_{i} \gets \val(\G^{\cdot,\strab}) $\label{line:si-u}
		\Until{$\ub_i(\initstate)-\lb_i(\initstate) \leq \varepsilon$}
	\end{algorithmic}
\end{algorithm}

\cref{alg:SI} shows how we can modify \cref{alg:MDP-view} to work with strategy iteration as the strategy recommender.
The key elements stay the same: in every iteration, the strategy recommender gives us a pair of strategies (\cref{line:si-infer}) which are then used to update the under- and over-approximations (Lines \ref{line:si-l} and \ref{line:si-u}).

\begin{lemma}
	Let $\Phi$ a reachability, safety, or mean payoff objective.
	Then \cref{alg:SI} is correct and terminates, i.e.\ for all $i$, we have $\lb_i \leq \val_{\G, \Phi} \leq \ub_i$, and for every $\varepsilon \geq 0$, there is $i$ with $\ub_i(\initstate) - \lb_i(\initstate) \leq \varepsilon$.
\end{lemma}
\begin{IEEEproof}
	The proof is analogous to the proof of \cref{thm:mdp-view}.
	It only remains to show that strategy iteration yields a strategy recommender.
	The sequence of Maximizer strategies is monotonically improving, see e.g.\ \cite{gandalf} for reachability (and by duality safety) and \cite{DBLP:conf/cdc/AkianCDG13} for mean payoff.
	Moreover, there are only finitely many strategies, and hence eventually $\straa$ is optimal and we compute the optimal strategy of Minimizer $\strab$ as the best response in \cref{line:si-mdp,line:si-infer}.
	Thus, strategy iteration is a stable strategy recommender.
\end{IEEEproof}

We comment on several interesting details:
Firstly, we deliberately exclude total reward, since to the best of our knowledge strategy iteration for total reward has not been considered so far.

%

%
%
Further, in \cref{line:si-l} we reuse the $x$ computed in \cref{line:si-mdp}.
Moreover, observe that by solving the MDP in \cref{line:si-mdp}, we typically directly obtain the optimal counter-strategy to $\straa$, i.e.\ $\strab$.
So when obtaining $\straa$ and $\strab$ from $x$, $\strab$ is exactly the globally optimal strategy in the MDP $\G^{\straa,\cdot}$; in contrast, the updated Maximizer strategy $\straa$ is only locally optimal against $\strab$.
Trying to use a global best response to $\strab$, i.e. solving the MDP $\G^{\cdot,\strab}$, can result in cyclic behaviour and non-convergence to the optimal strategies~\cite{condonAlgo}.
This is why the computation of $\ub_i$ does not affect the choice of the next strategies at all, but only serves to provide an upper bound.
This is also the only real difference to classic strategy iteration:
We additional compute the upper bound in \cref{line:si-u}.

Next, we can also consider the dual algorithm, which computes the sequence of Minimizer strategies.
Starting from an arbitrary Minimizer strategy $\strab$, we compute $x$ as $\val(\G^{\cdot,\strab})$ and thus get an upper bound $\ub_i$ on the value.
For the lower bound, we compute $\val(\G^{\straa,\cdot})$ with the current counter-strategy $\straa$ of Maximizer.

Finally, since our algorithm only complements standard SI with bounds on the precision, it does not affect the complexity of SI.
Thus, we know that it takes at most exponentially many iterations (as there are exponentially many MD strategies); and in the worst case, it can require exponentially many iterations~\cite{DBLP:journals/corr/abs-1106-0778}.
Independently, if we do not require an exact solution, our anytime bounds can lead to earlier termination.

\section{Computing the Staying Value} \label{app:MP-stayVal}

In this section, we describe how to compute the staying value.
We group reachability, safety and total reward objectives together in \appref{app:stayValOthers}, because for all these objectives the staying value can be obtained by graph analysis.
\appref{app:stayValMP} describes the computation for mean payoff.
Here, we differentiate precisely computing the staying value and approximating it.
The latter is more practical, since we can interleave computation of the staying value with the overall computation, in the spirit of \cite{DBLP:conf/cav/AshokCDKM17}.

\subsection{Reachability, Safety and Total Reward}\label{app:stayValOthers}


In this subsection, we briefly describe how to compute the staying value for all objectives we consider except mean payoff.
Recall that we compute the staying value for a given memoryless Minimizer strategy $\strab$, i.e.\ we consider the MDP $\G^{\cdot, \strab}$ restricted to an EC $E$.
(Or, dually, for a given Maximizer strategy.)
As such, this is a well known problem for all three objectives.
For reachability and safety, recall that we assumed target states to be absorbing, thus an EC either equals a target state or does not contain any at all.
As such, the staying values in these cases trivially are $0$ and $1$ for any non-target EC.
(Note that this logic can also be applied directly on $\G$.)

For total reward, recall that we assumed no state has infinite value \emph{in the game}.
Yet, it may still be the case that against a sub-optimal strategy $\strab$, Maximizer can actually achieve a value of infinity.
However, the computation still is rather straightforward:
Since $E$ is an EC in $\G^{\cdot, \strab}$, we can reach any state infinitely often.
Thus, if there exists a state with non-zero reward in $E$, we can definitely obtain infinite total reward.
Otherwise, i.e.\ all states in $E$ have $\statereward(s) = 0$, clearly the staying value is $0$.
In an MDP $\G^{\straa,\cdot}$ where Minimizer is playing, an EC has infinite staying value if Minimizer cannot avoid visiting a state with non-zero reward infinitely often.
This can be computed using standard graph analysis.
Note that such an EC with infinite staying value cannot be simple, since by assumption Minimizer has a strategy that exits every EC containing a state with non-zero reward (already in the game, i.e.\ even against an optimal Maximizer strategy).

\subsection{Mean Payoff} \label{app:stayValMP}

For mean payoff, the staying value of an EC $E$ can have a non-extremal value, i.e. not 0, 1 or $\infty$.
As before, the computation corresponds to solving the MDP $\G_{\restrict}^{\cdot, \strab}$.

If we want the precise value, we can obtain it by classical methods, such as linear programming or strategy iteration \cite[Chp.~8 \& 9]{DBLP:books/wi/Puterman94}.
However, solving all ECs precisely in each step is expensive; moreover, it is unnecessary when either (i)~$E$ was a wrong guess for a SEC or (ii)~full precision in the EC is not needed to achieve convergence, e.g.\ because exiting it is optimal.
We now describe how to modify \cref{alg:SG-view} to a more \enquote{on-demand}, purely VI style, i.e.\ to rely on successive approximations of the staying values.
This corresponds to the on-demand value iteration for mean payoff in MDP, see~\cite[Sec.~3.3]{DBLP:conf/cav/AshokCDKM17}.
We replace $\stayVal_{\G^{\cdot, \strab}, \Phi}(E,s)$ with an over-approximation $\upperStayVal_{\G^{\cdot, \strab}, \Phi}(E,s)$ in \cref{line2:defl-UM}, and, dually, use $\lowerStayVal_{\G^{\straa, \cdot}, \Phi}(E,s)$ in \cref{line2:infl-LM}.

We compute the approximations as follows:
\begin{itemize}
	\item \emph{Successively more precise value iteration}:
	We build on an idea from \cite[Sec.~3.3]{DBLP:conf/cav/AshokCDKM17}.
	We use value iteration to compute the value in the MDP.
	This allows us to stop before reaching full precision.
	Concretely, for a given EC $E$ in $\G^{\cdot, \strab}$ (or the dual for Minimizer), we have a precision $\varepsilon'$ (independent of the $\varepsilon$ of the overarching algorithm).
	We stop the local value iteration when precision $\varepsilon'$ is reached and can infer lower and upper bounds on the value of the restricted MDP;
	a more detailed description of this process follows in the proof of correctness below.
	The next time we consider the same EC, we decrease $\varepsilon'$, e.g.\ by halving it.
	On the one hand, this allows us to avoid spending time on ECs that were guessed wrongly, since we only solve them to precision $\varepsilon'$.
	On the other hand, since we have a convergent strategy recommender, from some point onward, all relevant ECs are found in every iteration of the main loop.
	Thus, the computation of staying values becomes more and more precise, eventually reaching the required precision.

	\item \emph{Early stopping}:
	We improve the previous idea with the additional insight that the staying value is relevant if and only if both players do not want to leave the EC.
	Thus, if the upper bound inferred from the value iteration is smaller than Maximizer's best exit, or dually the lower bound larger than Minimizer's best exit, we can also stop the value iteration.

	\item \emph{Initialization}:
	One could start the value iteration in the restricted MDP from a vector $x_0$ that assigns 0 to all states.
	However, the value iteration for mean payoff converges no matter which initial values are used.
	Thus, it is also correct and potentially a lot faster to start (i)~from the $x_i$ that the overarching algorithm has computed so far or (ii)~from the approximations that a previous computation of the staying value for the given EC has computed.
\end{itemize}
To summarize, we compute approximations $\upperStayVal_{\G^{\cdot, \strab}, \Phi}(E,s)$ and $\lowerStayVal_{\G^{\straa, \cdot}, \Phi}(E,s)$ as follows:
We run value iteration in the induced MDP restricted to the given end component, either until a required precision $\varepsilon'$ is reached or until the staying value is clearly irrelevant.
Also, we can re-use previous intermediate results, since the value iteration converges independently of the starting values.
We argue that this modification preserves correctness and convergence.
\begin{lemma}
	Modify \cref{alg:SG-view} as described, i.e.\ using $\upperStayVal_{\G^{\cdot, \strab}, \Phi}(E,s)$ in \cref{line2:defl-UM} and $\lowerStayVal_{\G^{\straa, \cdot}, \Phi}(E,s)$ in \cref{line2:infl-LM}; and computing these approximations of the staying value through successively more precise value iteration.

	This modified version is correct and terminates.
\end{lemma}
\begin{IEEEproof}
	The proof is the mainly the same as the proof of \cref{stm:algo_sg_correct}.
	We only address the differences.

	\emph{Correctness:}
%
%
	Correctness follows immediately from known results of mean payoff computation on MDP, which we recall briefly.
	In particular, we restate some claims of \cite[Sec.~8.5]{DBLP:books/wi/Puterman94}:
	Running value iteration according to the total reward Bellman equation (\cref{eq:bellmanTR}), we get a sequence of vectors $(x_i)_{i \in \Naturals}$.
	Let $\Delta_i \eqdef x_{i+1} - x_i$ be the difference vector, i.e.\ the reward gained in the $i+1$-th step.
	We can show that\footnote{
		Technically, we also need that the MDP is \emph{aperiodic}.
		For this, we employ a standard aperiodicity transformation from~\cite[Sec.~8.5.4]{DBLP:books/wi/Puterman94}.
	}
	\begin{equation}
		\min_{s \in \allstates} \Delta_i(s) \leq v \leq \max_{s \in \allstates} \Delta_i(s). \label{eq:mean_payoff_span_equation}
	\end{equation}
	Thus, when $\max_{s \in \allstates} \Delta_i(s) - \min_{s \in \allstates} \Delta_i(s) \leq \varepsilon'$, we know that we are $\varepsilon'$-close to the value $v$ and can stop value iteration.
	Moreover, $\min_{s \in \allstates} \Delta_i(s)$ is a lower bound on $v$ and $\max_{s \in \allstates} \Delta_i(s)$ an upper bound on $v$.
	In conclusion:
	Running value iteration as in~\cite[Sec.~8.5]{DBLP:books/wi/Puterman94} until $\varepsilon'$ precision is reached, we can return valid lower respectively upper bounds on the staying value in the given MDP.
	
	We address the two additional differences:
	For our \enquote{early stopping}, observe that we take the $\max$ in \cref{line2:defl-UM} and the $\min$ in \cref{line2:infl-LM}.
	This means that we stop computation early if and only if our approximation of the staying value certainly is smaller respectively larger than the exit value we compare to, meaning that further tightening the precision of the result has no influence on the algorithm anyway.
	Finally, for \enquote{restarting}, note that \cref{eq:mean_payoff_span_equation} is independent of the starting vector.
	
	\emph{Termination:}
	Since we have a convergent strategy recommender, eventually we always investigate the same ECs, namely those that are ECs under optimal strategies.
	Note that, as the strategies are optimal, all states of the EC have the same value \emph{in the SG}, and thus the ECs are SECs.
	Every time we approximate the staying value, we decrease the precision, and the limit of the sequence of our precisions is 0.
	The value iteration for mean payoff on the MDP is guaranteed to terminate for any precision requirement, thus, eventually the precision of the staying value is small enough that the overarching algorithm can terminate based on the computed staying values.
\end{IEEEproof}

\end{document}